\documentclass[10pt,twocolumn,letterpaper]{article}

\usepackage{3dv}      

\input{000_header.tex}

\threedvfinalcopy 
\def\threedvPaperID{xxx} 

\ifthreedvfinal\fi
\begin{document}

\title{A Benchmark and a Baseline for Robust Multi-view Depth Estimation}

\author{Philipp Schr\"oppel\\
University of Freiburg\\
\and
Jan Bechtold\\
Bosch\\
\and
Artemij Amiranashvili\\
University of Freiburg\\
\and
Thomas Brox\\
University of Freiburg\\
}

\maketitle
\thispagestyle{empty}

\begin{abstract}
Recent deep learning approaches for multi-view depth estimation are employed either in a depth-from-video or a multi-view stereo setting. Despite different settings, these approaches are technically similar: they correlate multiple \otherviews{} with a \keyview{} to estimate a depth map for the \keyview{}. In this work, we introduce the Robust Multi-view Depth Benchmark that is built upon a set of public datasets and allows evaluation in both settings on data from different domains. We evaluate recent approaches and find imbalanced performances across domains. Further, we consider a third setting where camera poses are available and the objective is to estimate the corresponding depth maps with their correct scale. We show that recent approaches do not generalize across datasets in this setting. This is because their cost volume output runs out of distribution. 
To resolve this, we present the \baselinename{} model for multi-view depth estimation, which is built upon existing components but employs a novel scale augmentation procedure. It can be applied for robust multi-view depth estimation, independent of the target data. We provide code for the proposed benchmark and baseline model at \small{\url{https://github.com/lmb-freiburg/robustmvd}}.

\end{abstract}
\section{Introduction} 

\begin{figure}[!ht]
\centering
\def\tmpwidth{0.3\linewidth}
\def\tmpwidthh{0.14\linewidth}
\begin{tabular}{@{}c@{\hspace{1mm}}c@{\hspace{1mm}}c@{\hspace{1mm}}c@{}}
    & \multicolumn{3}{c}{\textit{Zero-shot evaluation across domains and scales:}} \\
    \makecell{\rotatebox{90}{\kittishort{}}} &
    \makecell{\includegraphics[width=\tmpwidth]{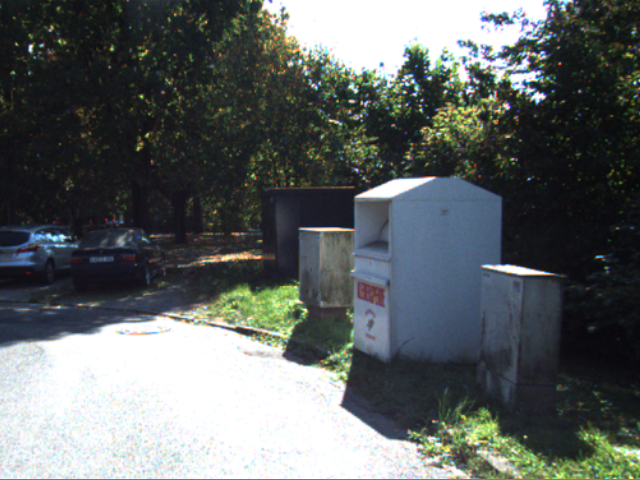}} &
    \makecell{\includegraphics[width=\tmpwidth]{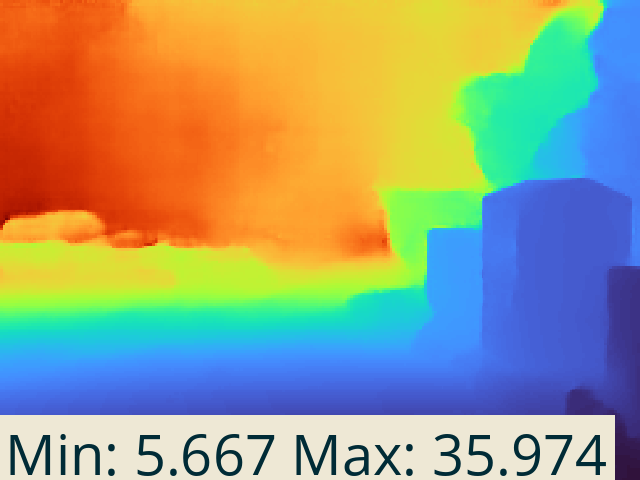}} &
    \makecell{\includegraphics[width=\tmpwidth]{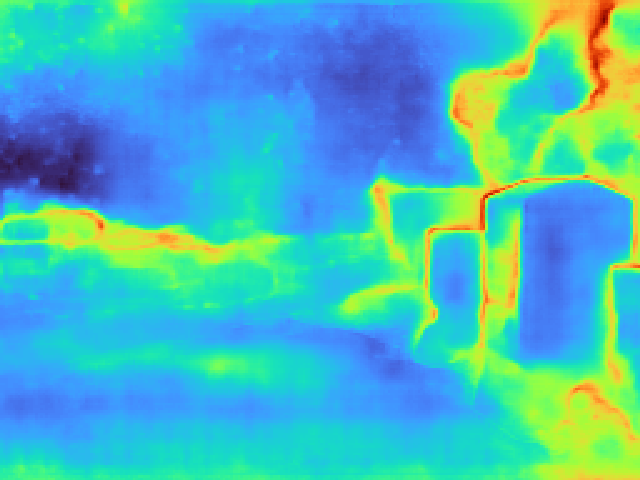}} \\
    \makecell{\rotatebox{90}{\scannetshort{}}} &
    \makecell{\includegraphics[width=\tmpwidth]{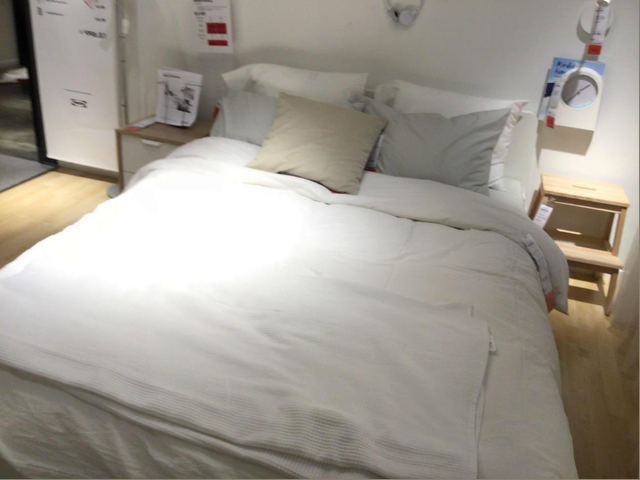}} &
    \makecell{\includegraphics[width=\tmpwidth]{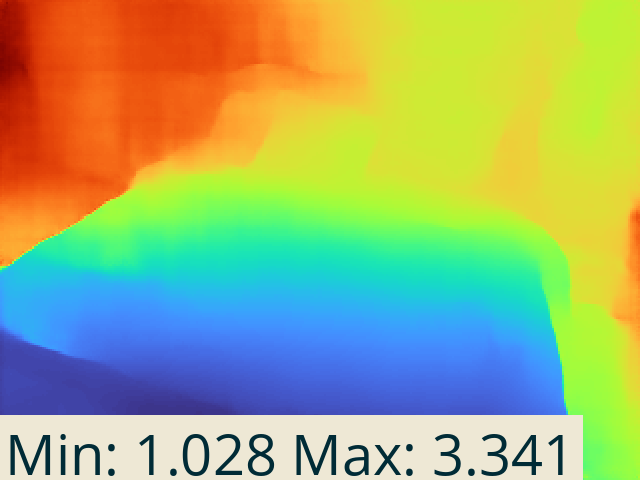}} &
    \makecell{\includegraphics[width=\tmpwidth]{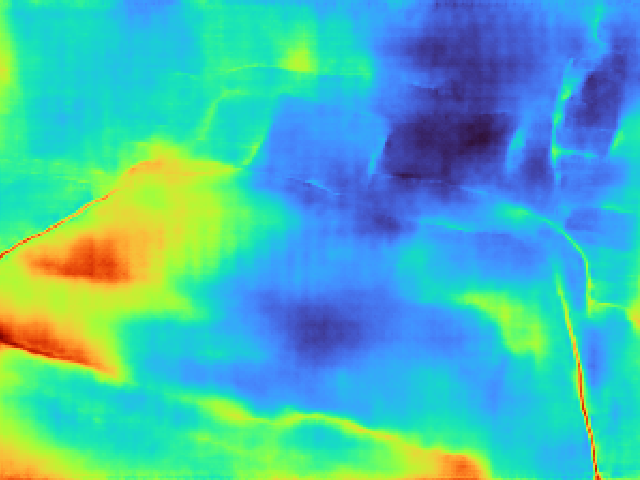}} \\
    \makecell{\rotatebox{90}{\ethdshort{}}} &
    \makecell{\includegraphics[width=\tmpwidth]{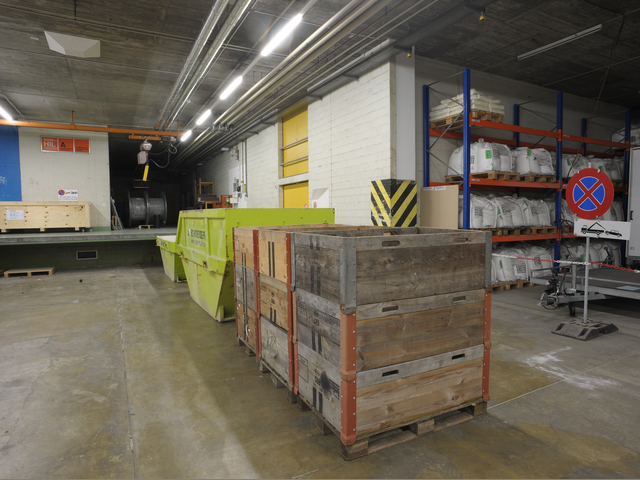}} &
    \makecell{\includegraphics[width=\tmpwidth]{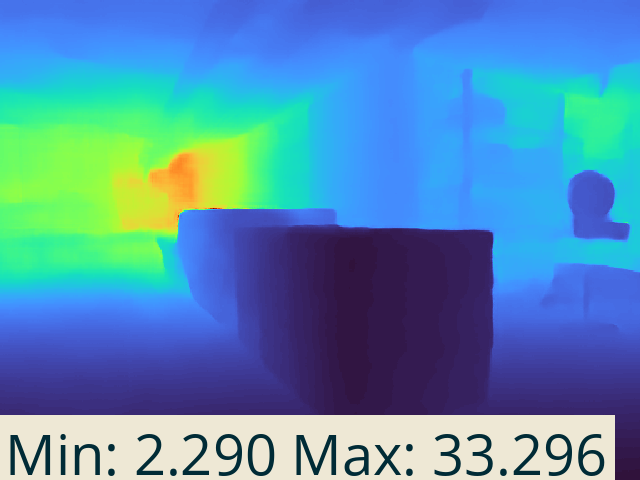}} &
    \makecell{\includegraphics[width=\tmpwidth]{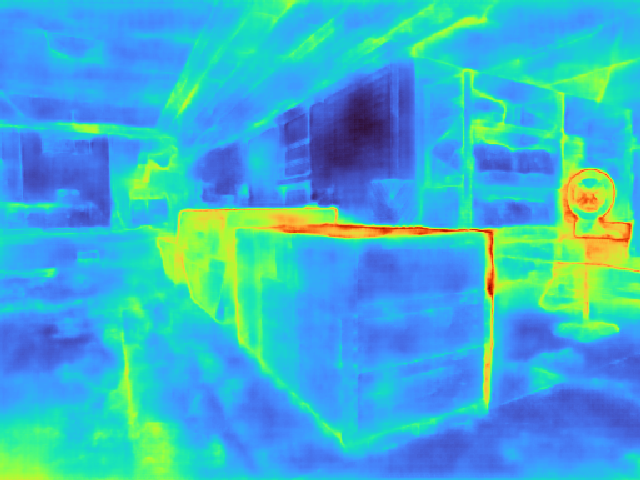}} \\
    \makecell{\rotatebox{90}{\tanksandtemplesshort{}}} &
    \makecell{\includegraphics[width=\tmpwidth]{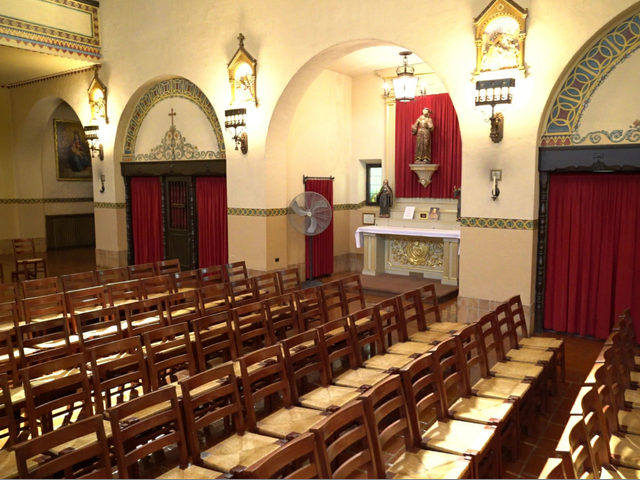}} &
    \makecell{\includegraphics[width=\tmpwidth]{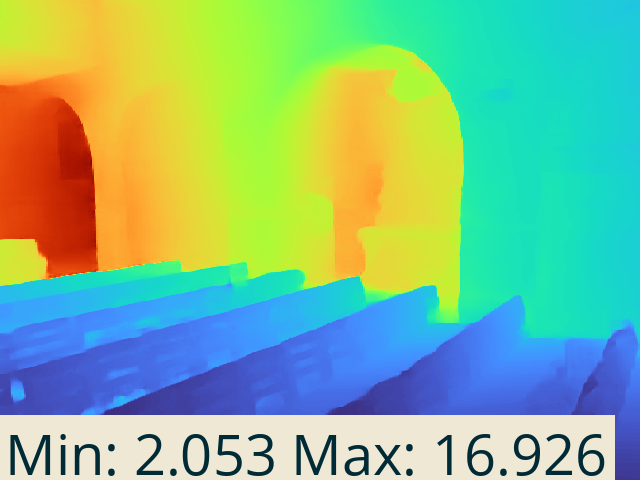}} &
    \makecell{\includegraphics[width=\tmpwidth]{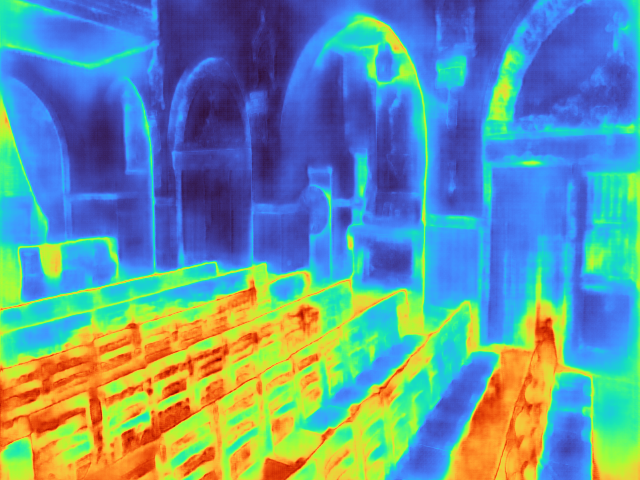}} \\
    \makecell{\rotatebox{90}{\dtu{}}} &
    \makecell{\includegraphics[width=\tmpwidth]{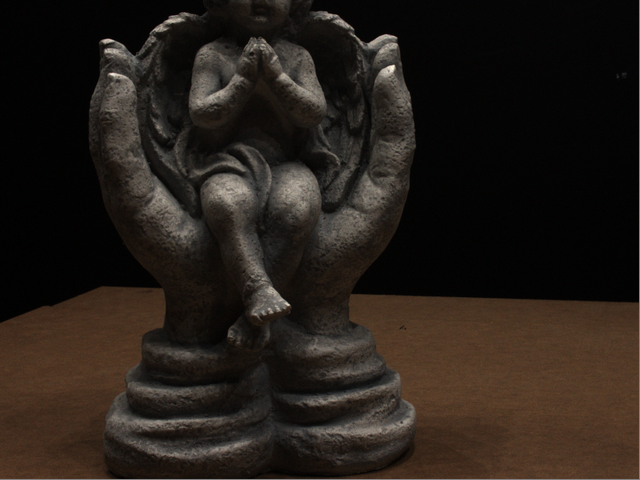}} &
    \makecell{\includegraphics[width=\tmpwidth]{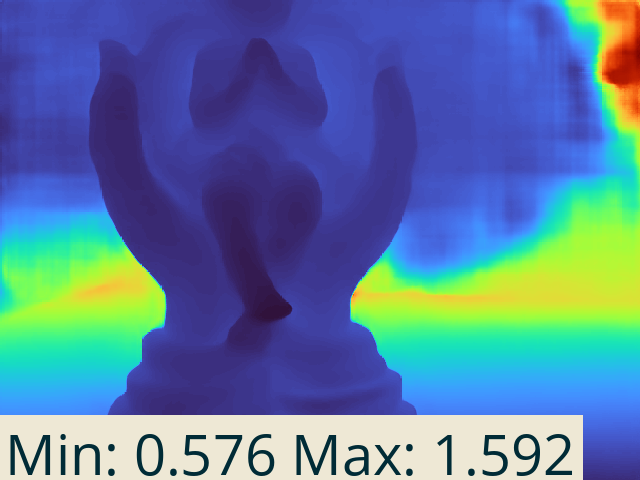}} &
    \makecell{\includegraphics[width=\tmpwidth]{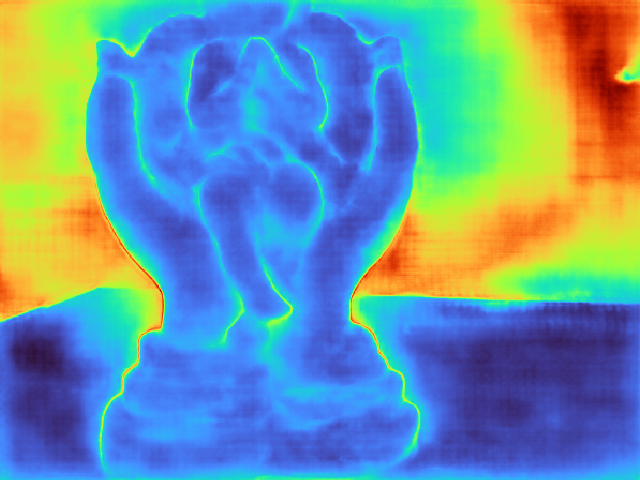}} \\
    & \Key{} Frame & Pred. Depth & Pred. Uncertainty
\end{tabular}
\caption{We introduce the Robust Multi-View Depth (MVD) Benchmark to evaluate multi-view depth estimation models regarding robust application on arbitrary real-world data. The benchmark comprises evaluation of depth and uncertainty estimates on multiple existing datasets from different domains and includes a setting with given camera poses and evaluation against absolute depth maps at real-world scale. 
We provide a baseline model that is built on existing components and generalizes well across domains and can be applied and extended for robust multi-view depth estimation. Predictions shown above are from the baseline model on benchmark data. \textcolor{myminturbo}{Purple} colors indicate small values, \textcolor{mymaxturbo}{red} colors large values, and the ranges are given in meters.}
\label{fig:teaser}
\end{figure}

Since the early days of computer vision, depth is reconstructed using the motion parallax between multiple views~\cite{Longuet1981, Kolmogorov2002, Schoenberger2016SfM, Schoenberger2016MVS}. The principle of motion parallax is generic.  It works the same in all domains, just like physics is the same everywhere in the world. Consequently, classical geometry-based approaches are not bound to a training distribution, but are agnostic to data from different domains. 

In recent years, approaches based on deep learning have emerged for multi-view depth estimation. They are employed either in a depth-from-video setting with images from a video with small and incremental but unknown camera motion~\cite{Ummenhofer2016,Zhou2018,Teed2020Deepv2d}, or a multi-view stereo setting with unstructured but calibrated image collections~\cite{Huang2018,Yao2018,Liu2019}.  Usually, at the core of these approaches are deep networks that correlate learned features from multiple images 
and learn to decode the obtained cost volume to an estimated depth map. This design, in principle, allows the network to base its estimates on the motion parallax, which should enable good generalization across domains and consistent predictions for different scene scales. However, approaches are often evaluated only on data similar to their training domain. Furthermore, evaluation is predominantly done only up to a relative scene scale: in depth-from-video, predictions are aligned to ground truth depths based on the median values; in multi-view stereo, models are supplied with minimum and maximum depth values and predict relative values within this range. 

In this work, we introduce a benchmark based on existing datasets to evaluate multi-view depth models regarding generalization across domains. 
Moreover, as specific cases like small camera motion, occlusions, or texture-less regions are potentially problematic, it is beneficial if a model comes with a measure of its uncertainty, which should be aligned with the depth prediction error.
In particular, the Robust Multi-view Depth Benchmark \begin{enumerate*}[label=(\arabic*)] \item evaluates estimated depth maps on data from different domains in a zero-shot fashion and \item evaluates uncertainties with the \ausename{} metric. Further, it \item includes evaluation in an absolute setting where ground truth camera poses are given to the model and evaluation is done against ground truth depths at their correct scale. As the scale is provided through the poses, evaluation is done without a given depth range and without alignment. This setting is relevant in practice, \eg in robotics or multi-camera setups where camera poses are known. \end{enumerate*}

We evaluate the depth and uncertainty estimates of recent models in their original relative depth-from-video or multi-view-stereo settings, as well as in the absolute depth estimation setting described above. We find that \begin{enumerate*}[label=(\arabic*)] \item almost all models have imbalanced performances across domains, \item uncertainties show only limited alignment with the prediction error, and \item models mostly perform well on a relative scale, but cannot be directly applied to estimate depths with their correct scale across datasets\end{enumerate*}. We attribute the problems at an absolute scale to out-of-distribution statistics in the correlation cost volume: depth-from-video models learn to use only the cost volume scores corresponding to absolute depth values seen during training; multi-view stereo models overfit to the cost volume distributions within the given minimum and maximum depth values and hence require a sufficiently accurate depth range to be known. 

The problems in depth estimation at absolute real-world scale limit practical application. 
To resolve this, we build a simple baseline model for robust cross-domain, scale-agnostic multi-view depth estimation. The model is mostly based on existing components, such as the DispNet architecture~\cite{Mayer2016} and trained on the \blendedmvs{} dataset~\cite{Yao2020} and a static version of the \flyingthings{} dataset~\cite{Mayer2016}. 
We only add scale augmentation as a new component to randomize across scales during training. 
This plain baseline achieves what the use of motion parallax promises: it generalizes across domains and scales. 
\section{Previous methods and benchmarks}

\vspace{-0.1em}
\paragraph{Depth-from-video}
In depth-from-video, depth maps are estimated from consecutive images of a video. Typically, it is assumed that the camera intrinsics are known, but not the camera motion. Hence, the task usually also comprises estimating the camera motion between images. DeMoN~\cite{Ummenhofer2016} was the first deep learning based approach for this task. DeMoN consists of a single network, which estimates depth and camera motion jointly from a pair of consecutive images. 
Later approaches are DeepTAM~\cite{Zhou2018} and DeepV2D~\cite{Teed2020Deepv2d}, which both process more than two frames, and estimate depth and camera motion with separate mapping and tracking modules, that are applied alternatingly. In such approaches, the mapping and tracking module typically overfit to the scene scale seen during training. Applying the models to scenes at a different scale requires aligning predictions to the scene scale based on additional information. Furthermore, our studies show that the mapping modules of such approaches do not generalize across scale,~\ie it is not generally possible to input ground truth camera poses at real-world scale and obtain absolute depth. We argue that this is a shortcoming, as the concept of mapping motion parallaxes to depths given camera motion is independent of the scale. 

\vspace{-1.4em}
\paragraph{Multi-view stereo}
In multi-view stereo, the task is to estimate the 3D geometry of an observed scene from an unstructured set of multiple images with known intrinsics and camera poses. 
Here, we focus on depth maps as a 3D geometry representation. DeepMVS~\cite{Huang2018} was the first deep network based approach for this task. DeepMVS brings the \keyview{} in correspondence with \otherviews{} with a correlation layer that samples patches from source images based on candidate depth values and compares them to patches from the key image. The resulting view-wise matching features are fused by max-pooling. MVSNet~\cite{Yao2018} takes a similar approach, but compares \otherviews{} and the \keyview{} in a learned feature space, and fuses multi-view information based on the variance across \otherviews{}. Many follow up works build upon this concept. R-MVSNet~\cite{Yao2018} reduces memory consumption by recurrent application.  CVP-MVSNet~\cite{Yang2020} and CAS-MVSNet~\cite{Gu2020} correlate in a coarse-to-fine fashion to reduce computational constraints and enable higher output resolutions. Vis-MVSNet~\cite{Zhang2020} improves fusion of multi-view information with a late-fusion strategy based on predicted uncertainties. Regarding different scene scales, all these approaches require the minimum and maximum depth value of the observed scene as input and predict depths relative to this range. Our studies show that these approaches have problems in a more general setting where ground truth poses are given, but the depth range of the observed scene is unknown.

\vspace{-1.4em}
\paragraph{Benchmarks and datasets}
Learned depth-from-video approaches are mostly evaluated on \kitti{}~\cite{Geiger2013,Uhrig2017} 
and \scannet{}~\cite{Dai2017}. KITTI is a benchmark suite for key tasks in vision-based autonomous driving, including depth estimation. ScanNet is a dataset for 3D scene understanding with annotated RGB-D videos of indoor scenes that were acquired at scale with an elaborate capturing framework. 
Learned multi-view stereo approaches are mostly evaluated on \dtu{}~\cite{Jensen2014,Aanaes2016}
, \ethd{}~\cite{Schoeps2017}
, and \tanksandtemples{}~\cite{Knapitsch2017}. 
\dtu{} consists of 80 scenes, each showing a tabletop object that was captured with a camera and a structured light scanner mounted on a robot arm. \tanksandtemples{} consists of real-world scenes that were captured indoors and outdoors with a high resolution video camera and an industrial laser scanner. 
Likewise, the \ethd{} high-resolution multi-view stereo benchmark consists of images of diverse indoor and outdoor scenes, captured with a high resolution DSLR camera and an industrial laser scanner. 
Training is often done on the same datasets, namely on \kitti{}, \scannet{}, and \dtu{}. Additionally, some approaches train on BlendedMVS~\cite{Yao2020}, which is designed specifically for large diversity to improve generalization. In this work, we additionally train on the \flyingthings{} dataset~\cite{Mayer2016}, which has been shown to enable good generalization in other matching-based tasks like disparity~\cite{Mayer2016} and optical flow estimation~\cite{ilg2017flownet,Teed2020Raft}.

\section{Robust Multi-view Depth Benchmark}
\label{sec:Experiments}

\paragraph{Key considerations}
In this work, we aim to evaluate multi-view depth models regarding robust depth estimation on arbitrary real-world data. To reflect this, we propose the Robust Multi-view Depth (MVD) Benchmark based on the following four key considerations:
\begin{enumerate}[itemsep=0.2ex,label=(\arabic*)]
    \item Depth estimation performance should be independent of the target domain. As a proxy, the benchmark defines test sets from diverse existing datasets. The training set is not defined, but must differ from test datasets. Evaluation is done in a zero-shot cross-dataset setting without fine-tuning. This simulates robustness to arbitrary, potentially unseen real-world data.
    \item The benchmark should be applicable to different multi-view depth estimation settings. 
    To this end, the benchmark allows different input modalities and optional alignment between predicted and ground truth depths. 
    \item Estimated uncertainty measures should be aligned with the depth estimation error. This is evaluated with the \ausename{} metric~\cite{Ilg2018}.
    \item The evaluation should not be affected by the selection of \otherviews{}. For this, a procedure to find and evaluate with a quasi-optimal set of source views is used.
\end{enumerate}
\enlargethispage{\baselineskip}

\vspace{-1em}
\paragraph{Relation to existing benchmarks} 
For multi-view stereo, multiple established benchmarks exist, \eg \dtu{}~\cite{Jensen2014,Aanaes2016}, \ethd{}~\cite{Schoeps2017}, and \tanksandtemples{}~\cite{Knapitsch2017}. We consider the proposed benchmark as complementary to these benchmarks. Existing multi-view stereo benchmarks evaluate 3D reconstruction performance on the basis of fused pointclouds. Complementarily, the proposed benchmark evaluates the generalization capabilities of learned models based on their typical raw outputs, namely depth maps and uncertainties. We encourage future works to evaluate 3D reconstruction performance on existing benchmarks, but additionally evaluate generalization capabilities on the \benchmarkname{}.
Depth-from-video models are usually trained and evaluated on existing datasets, \eg \kitti{} or \scannet{}. Usually the test sets are less diverse than in the proposed benchmark. We hence encourage future works on depth-from-video to evaluate generalization capabilities on the \benchmarkname{}. Depth estimation at absolute scale is usually not evaluated. However, we consider this setting as relevant in practice and encourage future work to evaluate in this setting on the \benchmarkname{}. 

\vspace{0.2em} In the following, we first describe the setup of the \benchmarkname{} in Sec.~\ref{sec:Setup}. We then present results of recent multi-view depth models, as well as the proposed \baselinename{} model on the benchmark in Sec.~\ref{sec:benchmark_results}. We provide details on the baseline model in Sec.~\ref{sec:Model}. 
\enlargethispage{\baselineskip}

\subsection{Setup}
\label{sec:Setup}

\paragraph{Test sets}
\label{sec:test_sets}

The test sets of the \benchmarkname{} are defined based on the \kitti~\cite{Uhrig2017}, \scannet{}~\cite{Dai2017}, \ethd{}~\cite{Schoeps2017}, \dtu{}~\cite{Jensen2014,Aanaes2016} and \tanksandtemples{}~\cite{Knapitsch2017} datasets, as they are common for multi-view stereo and depth-from-video evaluation and cover diverse domains and scene scales. 
\begin{table}[ht!]
\footnotesize
\centering
\resizebox{\columnwidth}{!}{ 
\begin{tabular}{|l||c|c|c|c|c|}

\hline
    \textbf{Test set}
    & \textbf{\kitti{}~\cite{Geiger2013, Uhrig2017}}
    & \textbf{\scannet{}~\cite{Dai2017}}
    & \textbf{\ethd{}~\cite{Schoeps2017}}
    & \textbf{\dtu{}~\cite{Jensen2014, Aanaes2016}}
    & \textbf{\tanksandtemplesshort{}~\cite{Knapitsch2017}}
    \\
    
\hline
\hline

    domain 
    & driving
    & indoor
    & in- \& outdoor
    & tabletop
    & in- \& outdoor
    \\

    setting 
    & DFV 
    & DFV 
    & MVS
    & MVS
    & MVS
    \\

    cam motion 
    & small
    & small
    & large
    & small
    & small
    \\

    scene scale 
    & \SI{2}-\SI{85}{\metre} 
    & \SI{0.2}-\SI{9}{\metre} 
    & \SI{0.3}-\SI{60}{\metre} 
    & \SI{0.4}-\SI{1.2}{\metre} 
    & \SI{1.1}-\SI{42}{\metre} 
    \\
    
\hline

    split 
    & test split from
    & test split from
    & orig. train 
    & val. split from 
    & orig. train 
    \\  
    
    based on  
    & Eigen~\etal~\cite{Eigen2014}
    & Tang and Tan~\cite{Tang2018}
    &   split
    & Yao~\etal~\cite{Yao2018}
    &   split
    \\ 
    
    
\hline

    full res. 
    & 1226x370
    & 640x480
    & 6048x4032
    & 1600x1200
    & 1962x1092
    \\

    
    \# samples
    & 93
    & 200
    & 104
    & 110
    & 69
    \\

\hline
\end{tabular}
}
\caption{\textbf{Test sets of the \benchmarkname{}} are based on \kitti{}, \scannet{}, \ethd{}, \dtu, and \tanksandtemples{} (\tanksandtemplesshort{}). These datasets are commen for depth-from-video (DFV) or multi-view stereo (MVS) and cover different domains and scene scales.
\label{tab:test_sets}
}
\end{table}

\vspace{-1.5em}
Each test set is a set of samples from the respective dataset. Each sample has input views $\sampleinput = \left( \view_{0}, \cdots, \view_{k} \right)$, consisting of a \keyview{} $\view_{0}$ and \otherviews{} $\view_{1,..,k}$, and (potentially sparse) ground truth depth values $\gtdepths$ for the \keyview{}. Each view $\view_i=(\image_i, \transformfromto{i}{0}, \intrinsic_i)$ consists of an image $\image_i$, a pose $\transformfromto{i}{0}$ relative to the \keyview{}, and intrinsics $\intrinsic_i$. 
The task is to estimate a dense depth map $\preddepthmap$ for the \keyview{} $\view_0$ from the input data. 
The test sets are chosen such that they are as comparable to existing data splits as possible. The test sets are deliberately rather small to speed up evaluation, but samples have been selected to cover a large diversity. 
An overview of the test sets is given in Tab.~\ref{tab:test_sets} and further details are provided in the Appendix.

\paragraph{Training set} The benchmark does not define a training set, as the objective is robustness to arbitrary real-world data independent of a specific training setup. However, it must be specified in case training data is used that overlaps with test datasets of the benchmark. 

\vspace{-1.0em}
\paragraph{Evaluation settings} The benchmark allows evaluation with different input modalities that are provided to the model and with an optional alignment between predicted and ground truth depths. The provided input modalities always include the images $\image_i$ and intrinsics $\intrinsic_i$ for each view and can optionally include the poses $\transformfromto{i}{0}$ and the ground truth depth range $(\gtdepthval_{min}, \gtdepthval_{max})$ with minimum and maximum ground truth depth values. To account for the scale-ambiguity of some models, predicted depth maps can optionally be aligned to the ground truth depth maps before computing the metrics, \eg based on the ratio of the median ground truth depth and the median predicted depth.

In literature, depth-from-video models are typically applied without poses and ground truth depth range and evaluated with alignment. Multi-view stereo models are typically applied with poses and ground truth depth range and evaluated without alignment.
Both settings evaluate depth estimations on a relative scale, \ie up to an unknown scale factor or within a given depth range. In contrast, the benchmark additionally evaluates depth estimation on an absolute scale. Here, the models are provided with poses but without depth range and the task is to estimate depth maps at absolute real-world scale. Evaluation is done without alignment.

\vspace{-1.0em}
\paragraph{Depth estimation metrics} 
Results are reported per test set for the \absrelname{} ($\absrel$) and the \threshIname{} ($\threshI$) with a threshold of $1.03$~\cite{Eigen2014,Uhrig2017}:
\begin{align}
    \absrel= & 100 \cdot \frac{1}{n}\sum\limits_{i=1}^n\frac{1}{m}\sum\limits_{j=1}^m\frac{\abs{\preddepthval_{i, j} - \gtdepthval_{i,j}}}{\gtdepthval_{i, j}} \\
    \threshI= & 100 \cdot \frac{1}{n}\sum\limits_{i=1}^n\frac{1}{m}\sum\limits_{j=1}^m [\max\left(\frac{\preddepthval_{i, j}}{\gtdepthval_{i, j}}, \frac{\gtdepthval_{i, j}}{\preddepthval_{i, j}}\right) < 1.03]
\end{align}
where $j$ indexes the $m$ pixels with valid ground truth depth, $i$ indexes the $n$ samples in a test set, and $[\cdot]$ denotes the Iverson bracket. The \absrelname{} indicates the average relative deviation of predicted depth values from ground truth depth values in percent. The \threshIname{} indicates the percentage of pixels with correct predictions, where a prediction is considered correct if it has an error below \SI{3}{\%}.  In addition to results on individual test sets, average metrics and model runtimes are reported over all test sets. 

Estimated depth maps are upsampled to the full resolution before computing the metrics. 
Additionally, to remove the effect of implausible outliers, depth estimates are clipped to a range of \SI{0.1}{\metre} to \SI{100}{\metre}. We conjecture that this is a reasonable range for real-world application. 

\vspace{-1.0em}
\paragraph{Uncertainty estimation metrics} 
\label{sec:uncertainty_estimation_metrics}
Results are reported with commonly used Sparsification Error Curves and the \ausename{} ($\ause$) metric~\cite{Ilg2018}. For the Sparsification Error Curves, the most erroneous pixels are gradually excluded from the error metric based on actual pixel errors (oracle uncertainty) versus estimated pixel uncertainties. The Sparsification Error Curve then is the difference of the oracle-based and uncertainty-based error reduction. The $\ause$ is the area under the Sparsification Error Curve. An $\ause$ of 0 is optimal and indicates perfect alignment between estimated uncertainties and actual errors. More details are provided in the Appendix.

\vspace{-0.5em}
\paragraph{Source view selection}
To factor out effects from the selection of \otherviews{} on the model performance, the benchmark finds and evaluates with a quasi-optimal set of \otherviews{} for each model. For a given sample, the model is run for all pairs $(\view_0, \view_i)$ of the \keyview{} and a single \otherview{} and the resulting Absolute Relative Errors are stored. The set of \otherviews{} is then grown incrementally by adding the \otherviews{} in the order of the stored Absolute Relative Errors. Results are reported for the set of \otherviews{} with the overall lowest \absrelname{}. Additionally, the \absrelname{} is plotted over the size of the \otherview{} set. 

\subsection{\benchmarkname{} results}
\label{sec:benchmark_results}

\begin{table*}[t!]
\footnotesize
\centering
\resizebox{\textwidth}{!}{
\setlength{\tabcolsep}{0.8mm}
\begin{tabular}{|l|c|c|c
|c >{\columncolor{bgcolor}}c
|c >{\columncolor{bgcolor}}c
|c >{\columncolor{bgcolor}}c
|c >{\columncolor{bgcolor}}c
|c >{\columncolor{bgcolor}}c
|c >{\columncolor{bgcolor}}c c
|}

\hline
    \textbf{Approach}
    & \textbf{\scriptsize{GT}}
    & \textbf{\scriptsize{GT}} 
    & \textbf{Align}
    & \multicolumn{2}{c|}{\textbf{\kittishort{}}}
    & \multicolumn{2}{c|}{\textbf{\scannetshort{}}}
    & \multicolumn{2}{c|}{\textbf{\ethdshort{}}}
    & \multicolumn{2}{c|}{\textbf{\dtushort{}}}
    & \multicolumn{2}{c|}{\textbf{\tanksandtemplesshort{}}}
    & \multicolumn{3}{c|}{\textbf{Average}}
    \\

    & \textbf{\scriptsize{Poses}}
    & \textbf{\scriptsize{Range}}
    &
    & $\absrel\downarrow$ & $\threshI\uparrow$
    & $\absrel\downarrow$ & $\threshI\uparrow$
    & $\absrel\downarrow$ & $\threshI\uparrow$
    & $\absrel\downarrow$ & $\threshI\uparrow$
    & $\absrel\downarrow$ & $\threshI\uparrow$
    & $\absrel\downarrow$ & $\threshI\uparrow$ & time [s] $\downarrow$
    \\
    \hline
    \hline

    \textbf{a)}
	& 
	& 
	& 
	& 
	& 
	& 
	& 
	& 
	& 
	& 
	& 
	& 
	& 
	& 
	& 
	&
    \\

    \textsc{Colmap}~\cite{Schoenberger2016MVS,Schoenberger2016SfM}
	& \my
	& \mn
	& \mn
	& \bestresult{12.0}
	& \bestresult{58.2}
	& \bestresult{14.6}
	& \bestresult{34.2}
	& \bestresult{16.4}
	& \bestresult{55.1}
	& \bestresult{0.7}
	& \bestresult{96.5}
	& \bestresult{2.7}
	& \bestresult{95.0}
	& \bestresult{9.3}
	& \bestresult{67.8}
	& $\approx\SI{3}{\minute}$
	\\ 

	\textsc{Colmap} \scriptsize{Dense}~\cite{Schoenberger2016MVS,Schoenberger2016SfM}
	& \my
	& \mn
	& \mn
	& 26.9
	& 52.7
	& 38.0
	& 22.5
	& 89.8
	& 23.2
	& 20.8
	& 69.3
	& 25.7
	& 76.4
	& 40.2
	& 48.8
	& $\approx\SI{3}{\minute}$
	\\ 
	
\hline
\hline

    \textbf{b)}
	& 
	& 
	& 
	& 
	& 
	& 
	& 
	& 
	& 
	& 
	& 
	& 
	& 
	& 
	& 
	&
    \\

	DeMoN~\cite{Ummenhofer2016}
	& \mn
	& \mn
	& $\Vert \vect t \Vert$
	& 15.5
	& 15.2
	& \bestresult{12.0}
	& \bestresult{21.0}
	& 17.4
	& 15.4
	& 21.8
	& 16.6
	& 13.0
	& 23.2
	& 16.0
	& 18.3
	& \bestresult{0.08}
	\\ 

	DeepV2D \scriptsize{\kitti{}}\normalsize{\cite{Teed2020Deepv2d}}
	& \mn
	& \mn
	& med
	& (\bestresult{3.1})
	& (\bestresult{74.9})
	& {23.7}
	& {11.1}
	& {27.1}
	& {10.1}
	& {24.8}
	& {8.1}
	& {34.1}
	& {9.1}
	& {22.6}
	& {22.7}
	& {2.07}
	\\ 

	DeepV2D \scriptsize{\scannet{}}\normalsize{~\cite{Teed2020Deepv2d}}
	& \mn
	& \mn
	& med
	& \bestresult{10.0}
	& \bestresult{36.2}
	& (\bestresult{4.4})
	& (\bestresult{54.8})
	& \bestresult{11.8}
	& \bestresult{29.3}
	& \bestresult{7.7}
	& \bestresult{33.0}
	& \bestresult{8.9}
	& \bestresult{46.4}
	& \bestresult{8.6}
	& \bestresult{39.9}
	& 3.57
	\\ 
	
\hline
\hline

    \textbf{c)}
	& 
	& 
	& 
	& 
	& 
	& 
	& 
	& 
	& 
	& 
	& 
	& 
	& 
	& 
	& 
	&
    \\

	MVSNet~\cite{Yao2018}
	& \my
	& \my
	& \mn
	& 22.7
	& 36.1
	& 24.6
	& 20.4
	& 35.4
	& 31.4
	& (\bestresult{1.8})
	& (86.0)
	& 8.3
	& 73.0
	& 18.6
	& 49.4
	& 0.07
	\\ 

	MVSNet \scriptsize{Inv. Depth}\normalsize{~\cite{Yao2018}}
	& \my
	& \my
	& \mn
	& 18.6
	& 30.7
	& 22.7
	& 20.9
	& 21.6
	& 35.6
	& (\bestresult{1.8})
	& (86.7)
	& 6.5
	& 74.6
	& 14.2
	& 49.7
	& 0.32
	\\ 

	CVP-MVSNet \normalsize{~\cite{Yang2020}}
	& \my
	& \my
	& \mn
	& {156.7}
	& {2.2}
	& {137.1}
	& {15.9}
	& {156.4}
	& {13.6}
	& ({4.0})
	& ({68.4})
	& {24.7}
	& {52.9}
	& {95.8}
	& {30.6}
	& {0.49}
	\\ 

	Vis-MVSNet~\cite{Zhang2020}
	& \my
	& \my
	& \mn
	& \bestresult{9.5}
	& \bestresult{55.4}
	& 8.9
	& 33.5
	& \bestresult{10.8}
	& \bestresult{43.3}
	& (\bestresult{1.8})
	& (\bestresult{87.4})
	& \bestresult{4.1}
	& \bestresult{87.2}
	& \bestresult{7.0}
	& \bestresult{61.4}
	& 0.70
	\\ 

	PatchmatchNet~\cite{Wang2020}
	& \my
	& \my
	& \mn
	& 10.8
	& 45.8
	& \bestresult{8.5}
	& \bestresult{35.3}
	& 19.1
	& 34.8
	& (2.1)
	& (82.8)
	& 4.8
	& 82.9
	& 9.1
	& 56.3
	& 0.28
	\\ 

	Fast-MVSNet~\cite{Yu2020}
	& \my
	& \my
	& \mn
	& 14.4
	& 37.1
	& 17.0
	& 24.6
	& 25.2
	& 32.0
	& (2.5)
	& (81.8)
	& 8.3
	& 68.6
	& 13.5
	& 48.8
	& 0.30
	\\ 

	MVS2D \scriptsize{\scannet{}}\normalsize{~\cite{Yang2022}}
	& \my
	& \my
	& \mn
	& 21.2
	& 8.7
	& (27.2)
	& (5.3)
	& 27.4
	& 4.8
	& 17.2
	& 9.8
	& 29.2
	& 4.4
	& 24.4
	& 6.6
	& \bestresult{0.04}
	\\ 

	MVS2D \scriptsize{\dtu{}}\normalsize{~\cite{Yang2022}}
	& \my
	& \my
	& \mn
	& 226.6
	& 0.7
	& 32.3
	& 11.1
	& 99.0
	& 11.6
	& (3.6)
	& (64.2)
	& 25.8
	& 28.0
	& 77.5
	& 23.1
	& 0.05
	\\ 
	
\hline
\hline

    \textbf{d)}
	& 
	& 
	& 
	& 
	& 
	& 
	& 
	& 
	& 
	& 
	& 
	& 
	& 
	& 
	& 
	&
    \\
    
	DeMoN~\cite{Ummenhofer2016}
	& \my
	& \mn
	& \mn
	& 16.7
	& 13.4
	& 75.0
	& 0.0
	& 19.0
	& 16.2
	& 23.7
	& 11.5
	& 17.6
	& 18.3
	& 30.4
	& 11.9
	& 0.08
	\\ 

	DeepTAM~\cite{Zhou2018}
	& \my
	& \mn
	& \mn
	& 68.7
	& 0.4
	& \trainedsimto{6.7}
	& \trainedsimto{39.7}
	& 20.4
	& 19.8
	& 58.0
	& 9.1
	& 40.0
	& 12.9
	& 38.8
	& 16.4
	& 0.85
	\\ 
	
	DeepV2D \scriptsize{\kitti{}}\normalsize{~\cite{Teed2020Deepv2d}}
	& \my
	& \mn
	& \mn
	& \trainedsimto{{20.4}}
	& \trainedsimto{{16.3}}
	& {25.8}
	& {8.1}
	& {30.1}
	& {9.4}
	& {24.6}
	& {8.2}
	& {38.5}
	& {9.6}
	& {27.9}
	& {10.3}
	& 1.43
	\\ 

	DeepV2D \scriptsize{\scannet{}}\normalsize{~\cite{Teed2020Deepv2d}}
	& \my
	& \mn
	& \mn
	& 61.9
	& 5.2
	& \trainedsimto{\bestresult{3.8}}
	& \trainedsimto{\bestresult{60.2}}
	& 18.7
	& 28.7
	& 9.2
	& 27.4
	& 33.5
	& 38.0
	& 25.4
	& 31.9
	& 2.15
	\\ 

	MVSNet\normalsize{~\cite{Yao2018}}
	& \my
	& \mn
	& \mn
	& 14.0
	& 35.8
	& 1568.0
	& 5.7
	& 507.7
	& 8.3
	& \trainedsimto{4429.1}
	& \trainedsimto{0.1}
	& 118.2
	& 50.7
	& 1327.4
	& 20.1
	& 0.15
	\\ 

	MVSNet \scriptsize{Inv. Depth}\normalsize{~\cite{Yao2018}}
	& \my
	& \mn
	& \mn
	& 29.6
	& 8.1
	& 65.2
	& 28.5
	& 60.3
	& 5.8
	& \trainedsimto{28.7}
	& \trainedsimto{48.9}
	& 51.4
	& 14.6
	& 47.0
	& 21.2
	& 0.28
	\\ 
	
	CVP-MVSNet\normalsize{~\cite{Yang2020}}
	& \my
	& \mn
	& \mn
	& 158.2
	& 1.2
	& 2289.0
	& 0.1
	& 1735.3
	& 1.2
	& \trainedsimto{8314.0}
	& \trainedsimto{0.0}
	& 415.9
	& 9.5
	& 2582.5
	& 2.4
	& 0.50
	\\ 

	Vis-MVSNet~\cite{Zhang2020}
	& \my
	& \mn
	& \mn
	& 10.3
	& \bestresult{54.4}
	& 84.9
	& 15.6
	& 51.5
	& 17.4
	& \trainedsimto{374.2}
	& \trainedsimto{1.7}
	& 21.1
	& 65.6
	& 108.4
	& 31.0
	& 0.82
	\\ 

	PatchmatchNet~\cite{Wang2020}
	& \my
	& \mn
	& \mn
	& 29.0
	& 16.3
	& 70.1
	& 16.7
	& 99.4
	& 3.5
	& (82.6)
	& (5.6)
	& 39.4
	& 19.3
	& 64.1
	& 12.3
	& 0.18
	\\ 

	Fast-MVSNet~\cite{Yu2020}
	& \my
	& \mn
	& \mn
	& 12.1
	& 37.4
	& 287.1
	& 9.4
	& 131.2
	& 9.6
	& (540.4)
	& (1.9)
	& 33.9
	& 47.2
	& 200.9
	& 21.1
	& 0.35
	\\ 

	MVS2D \scriptsize{\scannet{}}\normalsize{~\cite{Yang2022}}  
	& \my
	& \mn
	& \mn
	& 73.4
	& 0.0
	& (4.5)
	& (54.1)
	& 30.7
	& 14.4
	& 5.0
	& 57.9
	& 56.4
	& 11.1
	& 34.0
	& 27.5
	& \bestresult{0.05}
	\\ 

	MVS2D \scriptsize{\dtu{}}\normalsize{~\cite{Yang2022}}  
	& \my
	& \mn
	& \mn
	& 93.3
	& 0.0
	& 51.5
	& 1.6
	& 78.0
	& 0.0
	& (\bestresult{1.6})
	& (\bestresult{92.3})
	& 87.5
	& 0.0
	& 62.4
	& 18.8
	& 0.06
	\\ 

	\textbf{Robust MVD Baseline}
	& \my
	& \mn
	& \mn
	& \bestresult{7.1}
	& 41.9
	& \bestresult{7.4}
	& \bestresult{38.4}
	& \bestresult{9.0}
	& \bestresult{42.6}
	& \bestresult{2.7}
	& \bestresult{82.0}
	& \bestresult{5.0}
	& \bestresult{75.1}
	& \bestresult{6.3}
	& \bestresult{56.0}
	& 0.06
	\\ 

\hline
\end{tabular}
}
\caption{\textbf{Quantitative results for the evaluated multi-view depth models on the \benchmarkname{} with different evaluation settings:} \textbf{a)} Classical approaches. \textbf{b)} Evaluation without poses, without depth range, with alignment. This is the common setting in depth-from-video literature. \textbf{c)} Evaluation with poses, with depth range, without alignment. This is the common setting in multi-view stereo literature. \textbf{d)} Absolute scale evaluation with poses, without depth range, without alignment. Results are reported for the \absrelname{} ($\absrel$) and \threshIname{} ($\threshI$) with a threshold of $1.03$ on each test set and as averages across all test sets. Additionally, the average runtime in seconds of each model across all test sets is reported. All results are for the quasi-optimal selection of \otherviews{} of each model. (Parentheses) denote training on data from the same domain. \textbf{Bold} denotes best results.
\label{tab:eval_benchmark}
}
\end{table*}

\paragraph{Evaluated models} 
In this work, we evaluate the \textsc{Colmap}~\cite{Schoenberger2016MVS,Schoenberger2016SfM}, DeMoN~\cite{Ummenhofer2016}, DeepTAM~\cite{Zhou2018}, DeepV2D~\cite{Teed2020Deepv2d}, MVSNet~\cite{Yao2018}, CVP-MVSNet~\cite{Yang2020}, Vis-MVSNet~\cite{Zhang2020}, PatchmatchNet~\cite{Wang2020}, Fast-MVSNet~\cite{Yu2020}, and MVS2D~\cite{Yang2022} models on the proposed benchmark. This choice reflects seminal works that lay ground for later improvements, as well as works that represent the current state of the art. For all models, we use the original provided code and weights, except for MVSNet where we use the PyTorch implementation from Xiaoyang Guo, as it gave better performance than the original Tensorflow version. We additionally evaluate a MVSNet that we re-trained with plane sweep sampling in inverse depth space. For DeepV2D, we evaluate the \kitti{} and \scannet{} models. For MVS2D, we evaluate the \scannet{} and \dtu{} models. 
Note that we intentionally not re-train models on a specific uniform dataset, as the objective of the benchmark is generalization across diverse test sets, independent of the training data. 

\vspace{-1em}
\paragraph{Results }In Tab.~\ref{tab:eval_benchmark}, we report results of evaluated models on the proposed \benchmarkname{}. We report results up to a relative scale in the typical depth-from-video and multi-view stereo settings, as well as on an absolute scale. In the following, we discuss the results.

\fakeparagraph{Classical approaches} For a comparison to classical approaches, in Tab.~\ref{tab:eval_benchmark}a we report results of \textsc{Colmap}~\cite{Schoenberger2016MVS,Schoenberger2016SfM} on the benchmark. The results of applying \textsc{Colmap} with default parameters cannot be directly compared to those of learned models, as \textsc{Colmap} estimates depth maps at a lower density (\SI{54}{\%} in average) and we compute the metrics only for pixels with a valid prediction. We additionally report results for \textsc{Colmap} without filtering, which results in dense predictions but lower accuracy.

\fakeparagraph{Evaluation up to a relative scale}
In Tab.~\ref{tab:eval_benchmark}b and~\ref{tab:eval_benchmark}c, we report results in the typical depth-from-video and multi-view stereo settings up to a relative scale. 
It shows that all models perform significantly better on the training domain. 
\enlargethispage{\baselineskip}

\fakeparagraph{Evaluation on an absolute scale} In Tab.~\ref{tab:eval_benchmark}d, we provide results of the evaluated models in an absolute scale depth estimation setting. For DeepV2D and DeepTAM, we only use the mapping module and input ground truth poses. For models that require a given depth range, we assume an unknown depth range and provide a default range of \SI{0.2}{\metre} to \SI{100}{\metre}. This covers the range of all test sets and simulates real-world applications with no information except poses. 

In this setting, all evaluated models perform significantly worse. Depth-from-video models perform worse on datasets with a different depth range than the training data (\eg DeepV2D-\scannet{} on \kitti{}).  Multi-view stereo models perform worse on datasets where the depth differs from the given default depth range (\eg MVSNet on \dtu{}). Most evaluated models internally build and decode a cost volume, which is computed in a plane sweep stereo fashion by correlating \otherviews{} with the \keyview{} for specific (inverse) depth values. We attribute the performance decrease to out-of-distribution cost volume statistics. Depth-from-video models learn to use only the cost volume scores corresponding to absolute depth values seen during training. Multi-view stereo models overfit to the cost volume distribution within the provided depth range.

In practice, this means that existing depth-from-video models cannot be generally used with known ground truth camera poses. Multi-view stereo models in turn require a sufficiently accurate depth range of the observed scene to be known. Even though this depth range can be obtained by running structure-from-motion, this comes at the cost of increased runtime and complexity. 

\enlargethispage{\baselineskip}
The proposed \baselinename{} model shows consistent performance across all test sets. We conjecture that the model really learned to exploit multi-view cues that generalize across domains. Furthermore, the proposed scale augmentation enables absolute scale depth estimation independent of the scene scale. 

\begin{figure}[ht!]
\centering
\includegraphics[width=0.78\linewidth]{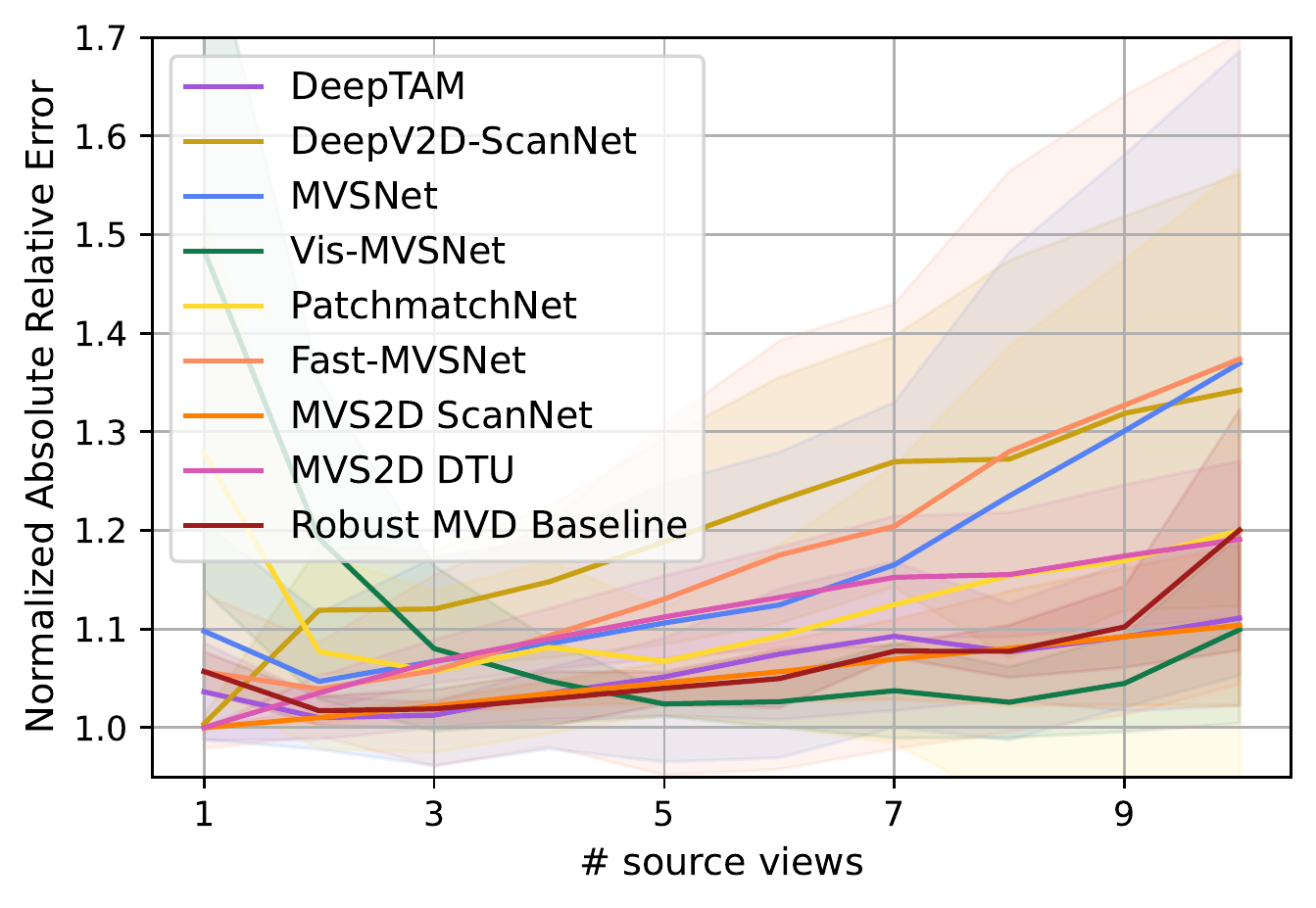}
\caption{\textbf{Effect of the number of \otherviews{} on the performance of evaluated models.} Each plot shows the average \absrelname{} across all test sets relative to the quasi-optimal performance of each model (Tab.~\ref{tab:eval_benchmark}). The shaded area indicates the standard deviation across test sets.}
\label{fig:eval_benchmark_num_views}
\end{figure}

\paragraph{Performances depending on \otherviews{}}
\label{sec:benchmark_results_num_views}
In Fig~\ref{fig:eval_benchmark_num_views}, we plot performances for different numbers of \otherviews{}. 
Additionally, model runtimes for different numbers of \otherviews{} are provided in the Appendix. 
For all models, we plot results in the respective setting that gives best average results according to Tab.~\ref{tab:eval_benchmark}.
In an ideal curve, the error would decrease with additional \otherviews{} and converge to a minimal value when more views do not contain additional information. 
The evaluation shows that multi-view fusion strategies of most models are suboptimal. 

\begin{figure}[h!]
\centering
\includegraphics[width=0.78\linewidth]{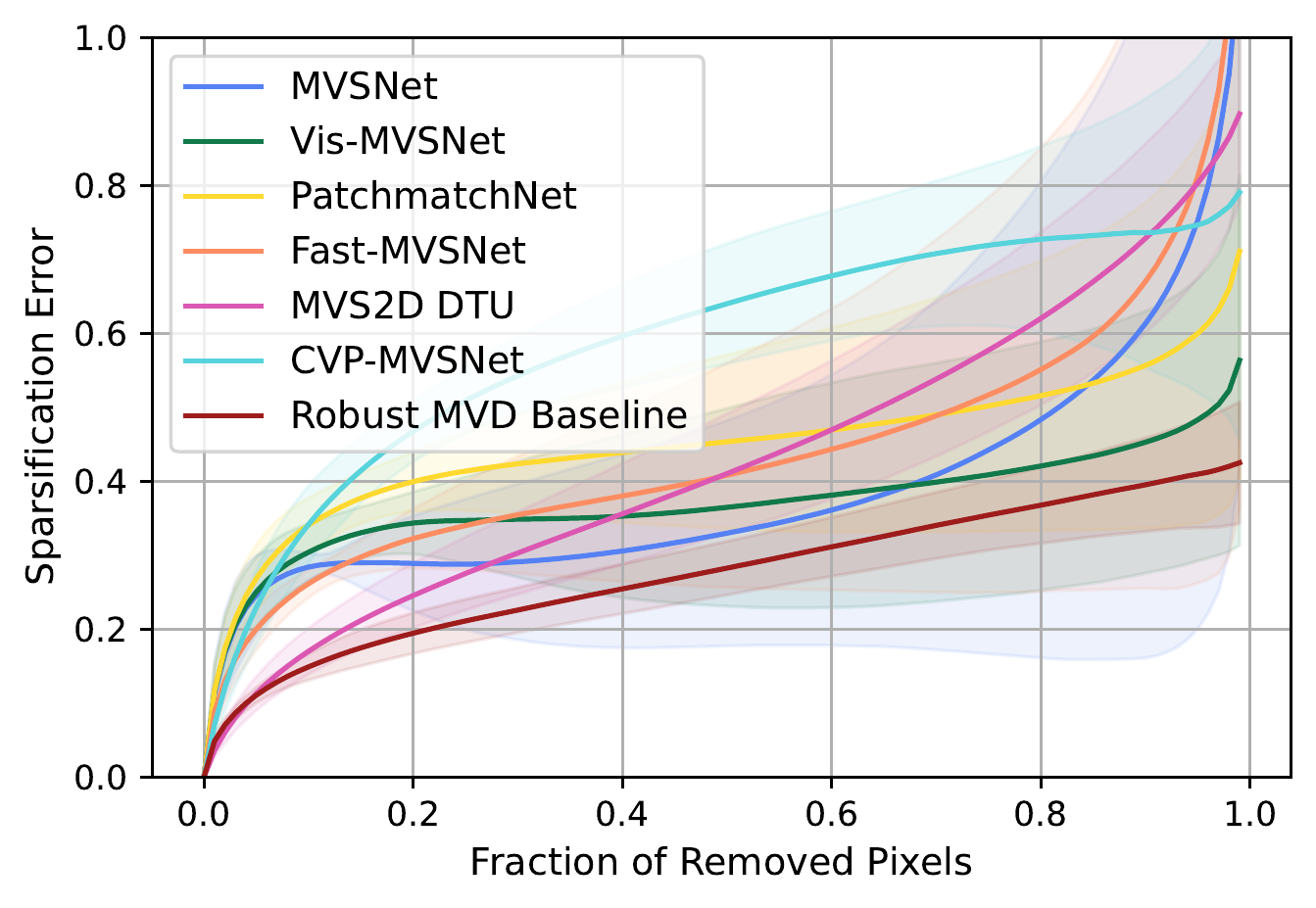}
\caption{\textbf{Evaluation of estimated uncertainty measures.} Lower curves indicate better alignment between estimated uncertainties and actual errors. The area under the curves is the \ausename{}, which is reported in Tab.~\ref{tab:eval_benchmark_uncertainty}.}
\label{fig:eval_benchmark_uncertainty}
\end{figure}

\paragraph{Uncertainty evaluation} In Fig.~\ref{fig:eval_benchmark_uncertainty}, we plot sparsification error curves for evaluated models that predict a measure of their depth prediction uncertainty. 
In Tab.~\ref{tab:eval_benchmark_uncertainty}, we report the corresponding \ausename{} metric. Again, we report results for each model in the respective setting that gives best average performance. 
The results of previous models show a suboptimal alignment between estimated uncertainties and errors, whereas the \baselinename{} model gives better uncertainties.  
\begin{table}[h!]
\footnotesize
\centering
\setlength{\tabcolsep}{0.2mm}
\begin{tabular}{|l
|c 
|c 
|c 
|c 
|c 
|c c 
|}

\hline
    \textbf{Approach}
    & \textbf{\kittishort{}}
    & \textbf{\small{\scannetshort{}}}
    & \textbf{\ethdshort{}}
    & \textbf{\dtushort{}}
    & \textbf{\tanksandtemplesshort{}}
    & \multicolumn{2}{c|}{\textbf{Average}}
    \\

    & 
    & 
    & 
    & 
    & 
    & $\absrel\downarrow$ & $\ause\downarrow$
    \\
    
    \hline
    \hline

	MVSNet \normalsize{~\cite{Yao2018}}
	& \bestresult{0.18}
	& 0.69
	& 0.35
	& 0.39
	& 0.32
	& 18.6
	& 0.39
	\\ 

	Vis-MVSNet~\cite{Zhang2020}
	& 0.28
	& 0.53
	& 0.37
	& \bestresult{0.27}
	& 0.39
	& 7.0
	& 0.37
	\\ 

	PatchmatchNet~\cite{Wang2020}
	& 0.47
	& 0.55
	& 0.52
	& 0.28
	& 0.40
	& 9.1
	& 0.45
	\\ 

	Fast-MVSNet~\cite{Yu2020}
	& 0.28
	& 0.73
	& 0.42
	& 0.29
	& 0.48
	& 13.5
	& 0.44
	\\ 

	MVS2D \scriptsize{\dtu{}}\normalsize{~\cite{Yang2022}}
	& 0.41
	& 0.50
	& 0.43
	& 0.31
	& 0.47
	& 77.5
	& 0.43
	\\ 
	
	CVP-MVSNet \normalsize{~\cite{Yang2020}}
	& {0.56}
	& {0.68}
	& {0.57}
	& {0.56}
	& {0.55}
	& {95.9}
	& {0.58}
	\\ 

	Robust MVD Baseline
	& 0.24
	& \bestresult{0.33}
	& \bestresult{0.28}
	& 0.28
	& \bestresult{0.24}
	& \bestresult{6.3}
	& \bestresult{0.27}
	\\ 

\hline
\end{tabular}
\caption{\textbf{Evaluation of estimated uncertainties} with the \ausename{} (AUSE). An AUSE of 0 means optimal alignment of uncertainties and errors.
\label{tab:eval_benchmark_uncertainty}
}
\end{table}

\section{\baselinename{}}
\label{sec:Model}

In the following, we describe the \baselinename{}, which is designed specifically as a baseline for robust depth estimation across domains and scene scales and can serve as baseline for evaluation on the proposed benchmark. The model is mostly based on existing components and we provide ablation studies for individual components in Tab.~\ref{tab:eval_model}. 

\begin{table*}[t!]
\footnotesize
\centering
\setlength{\tabcolsep}{2mm}
\begin{tabular}{|l
|c >{\columncolor{bgcolor}}c
|c >{\columncolor{bgcolor}}c
|c >{\columncolor{bgcolor}}c
|c >{\columncolor{bgcolor}}c
|c >{\columncolor{bgcolor}}c
|c >{\columncolor{bgcolor}}c c
|}

\hline
    \textbf{Approach}
    & \multicolumn{2}{c|}{\textbf{\kittishort{}}}
    & \multicolumn{2}{c|}{\textbf{\scannetshort{}}}
    & \multicolumn{2}{c|}{\textbf{\ethdshort{}}}
    & \multicolumn{2}{c|}{\textbf{\dtushort{}}}
    & \multicolumn{2}{c|}{\textbf{\tanksandtemplesshort{}}}
    & \multicolumn{3}{c|}{\textbf{Average}}
    \\

    & $\absrel\downarrow$ & $\threshI\uparrow$
    & $\absrel\downarrow$ & $\threshI\uparrow$
    & $\absrel\downarrow$ & $\threshI\uparrow$
    & $\absrel\downarrow$ & $\threshI\uparrow$
    & $\absrel\downarrow$ & $\threshI\uparrow$
    & $\absrel\downarrow$ & $\threshI\uparrow$ & time [ms] $\downarrow$
    \\
    \hline
    \hline

    \textbf{a) Scale augmentation}
	& 
	& 
	& 
	& 
	& 
	& 
	& 
	& 
	& 
	& 
	& 
	& 
	&
    \\

	No scale augmentation
	& 18.5
	& 19.6
	& 81.9
	& 8.9
	& 42.0
	& 15.8
	& 804.2
	& 0.0
	& 17.9
	& 43.2
	& 192.9
	& 17.5
	& 32.1
	\\ 

	With scale augmentation
	& \bestresult{15.2}
	& \bestresult{21.7}
	& \bestresult{8.5}
	& \bestresult{31.5}
	& \bestresult{19.4}
	& \bestresult{22.7}
	& \bestresult{5.7}
	& \bestresult{49.5}
	& \bestresult{14.5}
	& \bestresult{50.5}
	& \bestresult{12.7}
	& \bestresult{35.2}
	& {32.7}
	\\ 
	
    \hline
    \hline

    \textbf{b) Training data}
	& 
	& 
	& 
	& 
	& 
	& 
	& 
	& 
	& 
	& 
	& 
	& 
	&
    \\

	\staticthingsshort{}
	& 15.2
	& 21.7
	& \bestresult{8.5}
	& 31.5
	& 19.4
	& 22.7
	& 5.7
	& 49.5
	& 14.5
	& 50.5
	& 12.7
	& 35.2
	& {32.7}
	\\ 

	\blendedmvsshort{}
	& 11.1
	& 27.3
	& 9.3
	& 29.5
	& \bestresult{12.4}
	& \bestresult{31.6}
	& \bestresult{4.6}
	& \bestresult{62.9}
	& \bestresult{9.4}
	& 52.9
	& \bestresult{9.4}
	& 40.8
	& 34.7
	\\ 

	\staticthingsshort{}+\blendedmvsshort{}
	& \bestresult{10.2}
	& \bestresult{27.7}
	& 8.7
	& \bestresult{31.7}
	& 14.6
	& 30.7
	& 4.7
	& 61.4
	& 11.3
	& \bestresult{57.3}
	& 9.9
	& \bestresult{41.8}
	& 33.6
	\\ 

	
\hline
\hline

    \textbf{c) Model architecture}
	& 
	& 
	& 
	& 
	& 
	& 
	& 
	& 
	& 
	& 
	& 
	& 
	&
    \\

	MVSNet architecture
	& 11.3
	& \bestresult{29.7}
	& 15.2
	& 23.4
	& 36.8
	& 25.9
	& 123.8
	& 48.8
	& \bestresult{10.8}
	& \bestresult{60.4}
	& 39.6
	& 37.6
	& 193.2
	\\ 

	DispNet architecture
	& \bestresult{10.2}
	& 27.7
	& \bestresult{8.7}
	& \bestresult{31.7}
	& \bestresult{14.6}
	& \bestresult{30.7}
	& \bestresult{4.7}
	& \bestresult{61.4}
	& 11.3
	& 57.3
	& \bestresult{9.9}
	& \bestresult{41.8}
	& {33.6}
	\\ 
	
    \hline
    \hline

    \textbf{d) Uncertainty estimation}
	& 
	& 
	& 
	& 
	& 
	& 
	& 
	& 
	& 
	& 
	& 
	& 
	&
    \\
    
	Deterministic
	& 10.2
	& 27.7
	& 8.7
	& 31.7
	& 14.6
	& 30.7
	& 4.7
	& 61.4
	& 11.3
	& 57.3
	& 9.9
	& 41.8
	& 33.6
	\\ 

	Laplace distribution
	& \bestresult{9.3}
	& \bestresult{31.9}
	& \bestresult{8.2}
	& \bestresult{35.0}
	& \bestresult{11.7}
	& \bestresult{38.1}
	& \bestresult{3.4}
	& \bestresult{76.6}
	& \bestresult{9.1}
	& \bestresult{63.7}
	& \bestresult{8.4}
	& \bestresult{49.1}
	& {35.6}
	\\ 



	
\hline
\hline

    \textbf{e) Multi-view fusion}
	& 
	& 
	& 
	& 
	& 
	& 
	& 
	& 
	& 
	& 
	& 
	& 
	&
    \\

	1 source view
	& 9.3
	& 31.9
	& 8.2
	& 35.0
	& 11.7
	& 38.1
	& 3.4
	& 76.6
	& 9.1
	& 63.7
	& 8.4
	& 49.1
	& {35.6}
	\\ 

	Averaging
	& 6.7
	& 40.1
	& 7.5
	& 38.5
	& 9.7
	& 39.9
	& 3.0
	& 79.6
	& 6.0
	& 74.2
	& 6.6
	& 54.5
	& 58.6
	\\ 

	Learned view weights
	& \bestresult{6.6}
	& \bestresult{42.0}
	& \bestresult{7.4}
	& \bestresult{38.7}
	& 9.2
	& \bestresult{42.9}
	& 2.9
	& 80.6
	& 7.6
	& \bestresult{76.0}
	& 6.8
	& \bestresult{56.0}
	& 61.7
	\\ 

	Learned view weights + Eraser
	& 7.1
	& 41.9
	& \bestresult{7.4}
	& 38.4
	& \bestresult{9.0}
	& 42.6
	& \bestresult{2.7}
	& \bestresult{82.0}
	& \bestresult{5.0}
	& 75.1
	& \bestresult{6.3}
	& \bestresult{56.0}
	& 59.2
	\\ 
	
\hline
\end{tabular}
\caption{\textbf{Ablation studies for the \baselinename{} model}. All results are for the absolute scale depth estimation setting (Tab.~\ref{tab:eval_benchmark}d). 
\textbf{a)} Scale augmentation is essential for generalization across scene scales. \textbf{b)} Joint training on \staticthings{} and \blendedmvs{} gives best performance. \textbf{c)} A DispNet architecture performs better than a MVSNet architecture. \textbf{d)} Predicting parameters of a Laplace distribution instead of point estimates improves performance. \textbf{e)} Multi-view fusion via weighted averaging with learned weights work slightly better than simple averaging. The last model is the \baselinename{} model.
\label{tab:eval_model}
}
\end{table*}

\subsection{Model architecture} 
\label{sec:model_architecture}

\begin{figure}[h!]
\centering
\includegraphics[width=0.99\linewidth]{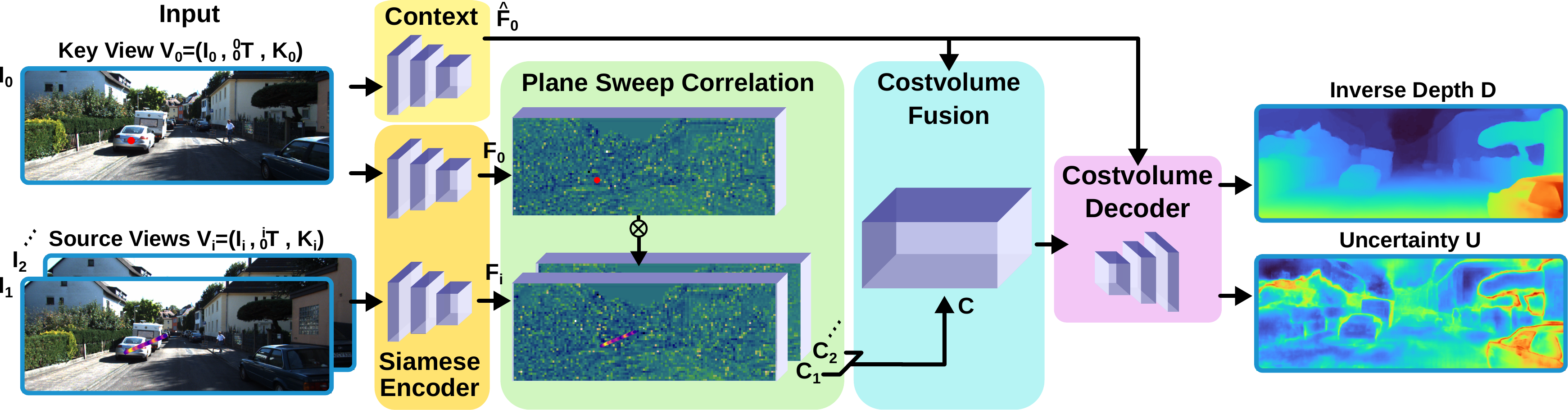}
\caption{\textbf{\baselinename{} model architecture}. The model consists of a siamese encoder network, a plane sweep correlation that correlates \otherviews{} with the \keyview{}, a fusion module, and a decoder that predicts inverse depths and uncertainties.}
\label{fig:method}
\end{figure}

The \baselinename{} model builds on the simple DispNet~\cite{Mayer2016} network architecture, but is adapted  to the given multi-view setting with non-rectified images. 
More specifically, as illustrated in Fig.~\ref{fig:method}, and using the notation defined in Sec.~\ref{sec:Setup}, the model architecture is structured as follows:
\begin{enumerate*}[label=(\arabic*)] 
\item a siamese encoder network $\encoder$ that maps input images $\image_i$ to feature maps, $\featuremap_i = \encoder\left(\image_i\right)$, 
\item a correlation layer that correlates \keyview{} features $\feature_0$ 
with \otherview{} features $\feature_i$ in a plane sweep fashion, resulting in view-wise cost volumes $\costvolume_{1,..,k}$, 
\item a context encoder network $\ctx$ that maps the \key{} image to features $\ctxfeaturemap_0 = \ctx\left(I_0\right)$ that are used to decode cost volumes,
\item a fusion module $\fusion$ that fuses the cost volumes from multiple \otherviews{} to a fused representation $\costvolume = \fusion(\costvolume_{1,..,k}, \ctxfeaturemap_0)$ via weighted averaging with learned weights, and
\item a 2D convolutional cost volume decoder network $(\predinvdepthmap, \uncertaintymap)=\decoder(\costvolume,\ctxfeaturemap_0)$ that maps the fused cost volume to an output inverse depth map $\predinvdepthmap$, and an uncertainty map $\uncertaintymap$\end{enumerate*}. The inverse depth map $\predinvdepthmap$ holds predicted inverted depth values $\predinvdepthval=1/\preddepthval$ for every \keyview{} pixel. The plane sweep correlation has been shown to work well in other multi-view depth architectures~\cite{Teed2020Deepv2d,Zhou2018,Yao2018,Liu2019,Huang2018} and is explained in the Appendix.

In the first experiments, we apply the base model in dual-view mode, using only a single \otherview{}. This factors out effects from the multi-view cost volume fusion and allows for an isolated evaluation of effects from data augmentation, training dataset, model architecture, and uncertainty estimation. Following this, we evaluate different strategies for fusing multi-view information. In Tab.~\ref{tab:eval_model}c, we compare the DispNet architecture with a MVSNet architecture. 

\enlargethispage{\baselineskip}

\subsection{Data augmentation} 
\begin{figure}[h!]
\centering
\includegraphics[width=0.99\linewidth]{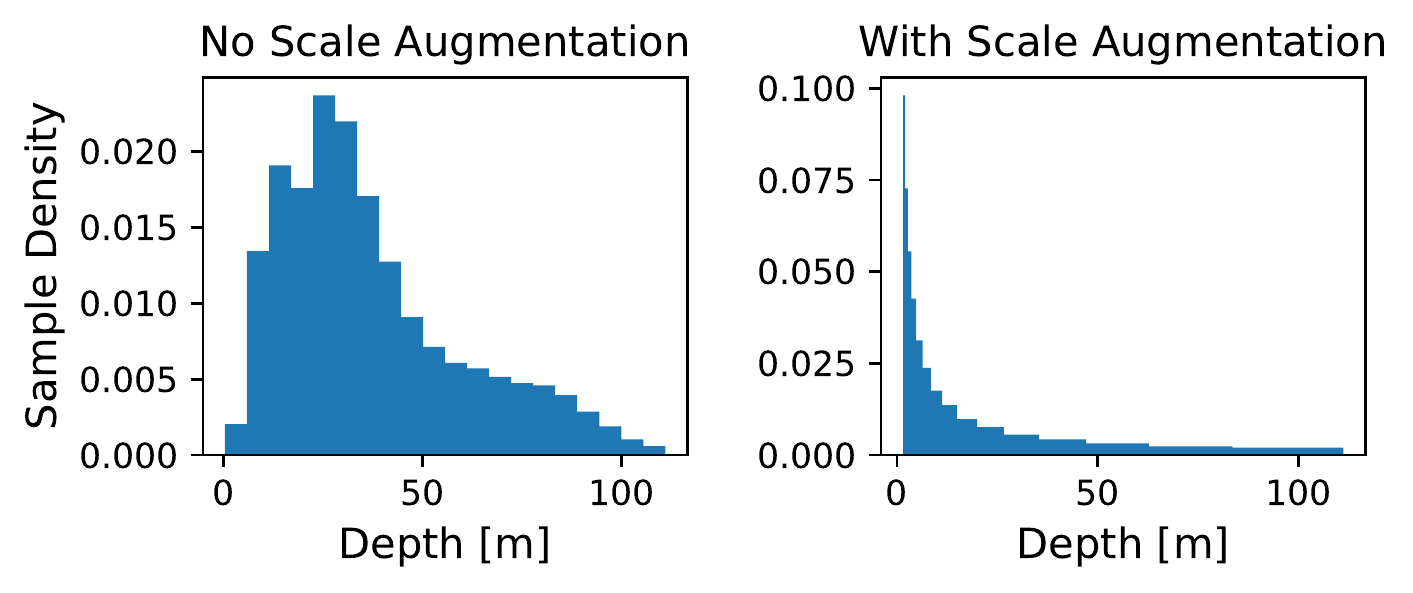}
    \caption{Effect of scale augmentation on the distribution of depth values seen during training on \staticthings{}. With scale augmentation, smaller depth values are sampled at a higher density. All bins in the "With Scale Augmentation" histogram cover the same area. This gives consistent performance across scene scales.}
\label{fig:scale_aug_hists}
\end{figure}

\begin{figure}[h!]
        \centering
        \subfloat[]{\includegraphics[width=0.24\linewidth]{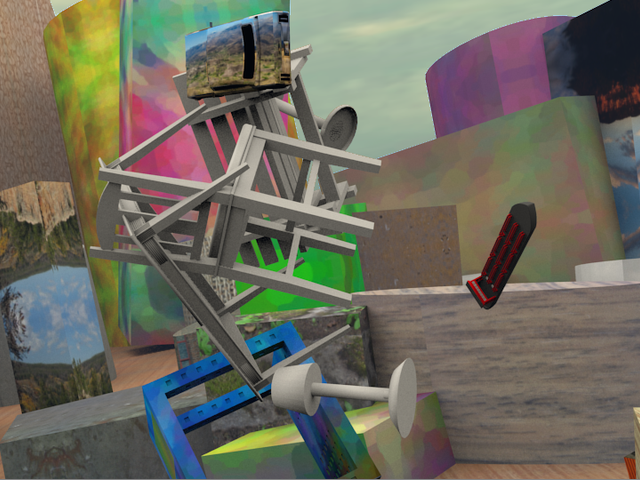}\label{subfig:DataAndAug-a} }
        \subfloat[]{
        \includegraphics[width=0.24\linewidth]{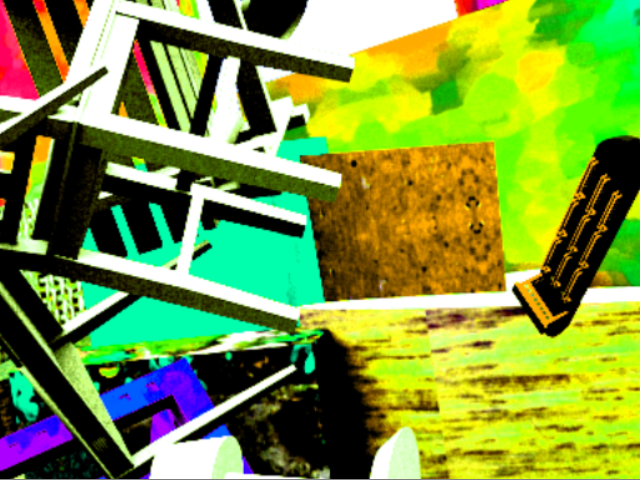}\label{subfig:DataAndAug-b} }
        \subfloat[]{
        \includegraphics[width=0.24\linewidth]{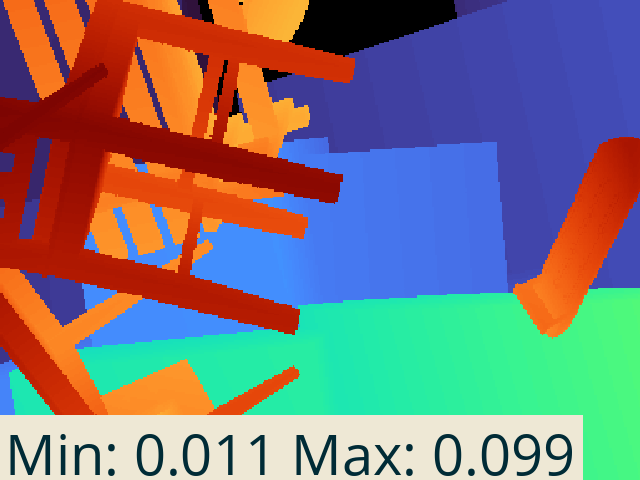}\label{subfig:DataAndAug-c} }
        \subfloat[]{
        \includegraphics[width=0.24\linewidth]{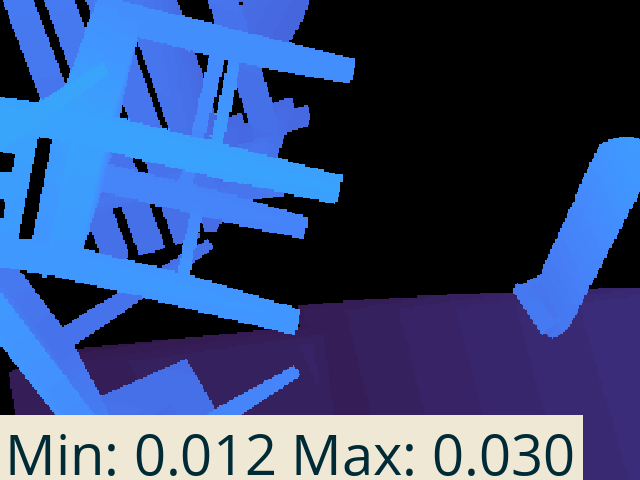}\label{subfig:DataAndAug-d}}
        \caption{\textbf{Training data and augmentation}: \protect\subref{subfig:DataAndAug-a} \keyview{} image $\image_0$ of a \staticthings{} training sample,  \protect\subref{subfig:DataAndAug-b} augmented \keyview{} image, \protect\subref{subfig:DataAndAug-c} ground truth inverse depth $\gtinvdepthmap$, and \protect\subref{subfig:DataAndAug-d} $\gtinvdepthmap$ after scale augmentation with a randomly sampled scaling factor of 3.27. Translations $\translationfromto{i}{0}$ to \otherviews{} are scaled with the same factor.}
        \label{fig:DataAndAug}
\end{figure}

Standard photometric and spatial augmentations are applied uniformly to all views. Additionally, to prevent the model from overfitting to the depth distribution of the training data, we introduce a novel data augmentation strategy that we term scale augmentation. Scale augmentation re-scales ground truth translations $\translationfromto{i}{0}$ during training before feeding them to the model. Likewise, the ground truth inverse depth map $\gtinvdepthmap$ is scaled with the inverse scaling factor. Inverse depth values outside the range $[\SI{0.009}{\metre^{-1}}, \SI{2.75}{\metre^{-1}}]$ are masked. To choose the scaling factor, a histogram of the depth values that were seen during previous training iterations, is maintained. The size of the histogram bins increases logarithmically, as consistent performance across the full depth range requires training for smaller depth values at a finer resolution. This is illustrated by Fig.~\ref{fig:scale_aug_hists}.
Scaling factors are then computed as the ratio of the depth label of the histogram bin with the lowest count and the median ground truth depth value of the current sample. Fig.~\ref{fig:DataAndAug} shows effects of the data augmentation on an exemplary sample. As shown by the results in Tab.~\ref{tab:eval_model}a, scale augmentation is a key component for enabling the model to generalize across different scene scales. 
\enlargethispage{\baselineskip}




\subsection{Training data} 
The \baselinename{} model is jointly trained on a static version of the existing \flyingthings{} dataset~\cite{Mayer2016}, that we term \staticthings{} (see Fig.~\ref{fig:DataAndAug}), and on the existing \blendedmvs{}~\cite{Yao2020} dataset. \staticthings{} is similar to \flyingthings{}: it contains 2250 train and 600 test sequences with 10 frames per sequence, showing randomly placed ShapeNet objects in front of random Flickr backgrounds. However, in \staticthings{}, all objects are static, and only the camera moves.  
The advantage of using this randomized synthetic dataset is that it reduces the possibility of a model to overfit to domain-specific priors. In Tab.~\ref{tab:eval_model}b, we compare joint training on \staticthings{} and \blendedmvs{} against training on a single dataset. Joint training performs quantitatively on par with training solely on \blendedmvs{}, but results in more accurate object boundaries, as shown in Fig.~\ref{fig:TrainData}. 
Further training details are provided in the Appendix.
\begin{figure}[h!]
        \centering
        \subfloat[]{\includegraphics[width=0.49\linewidth]{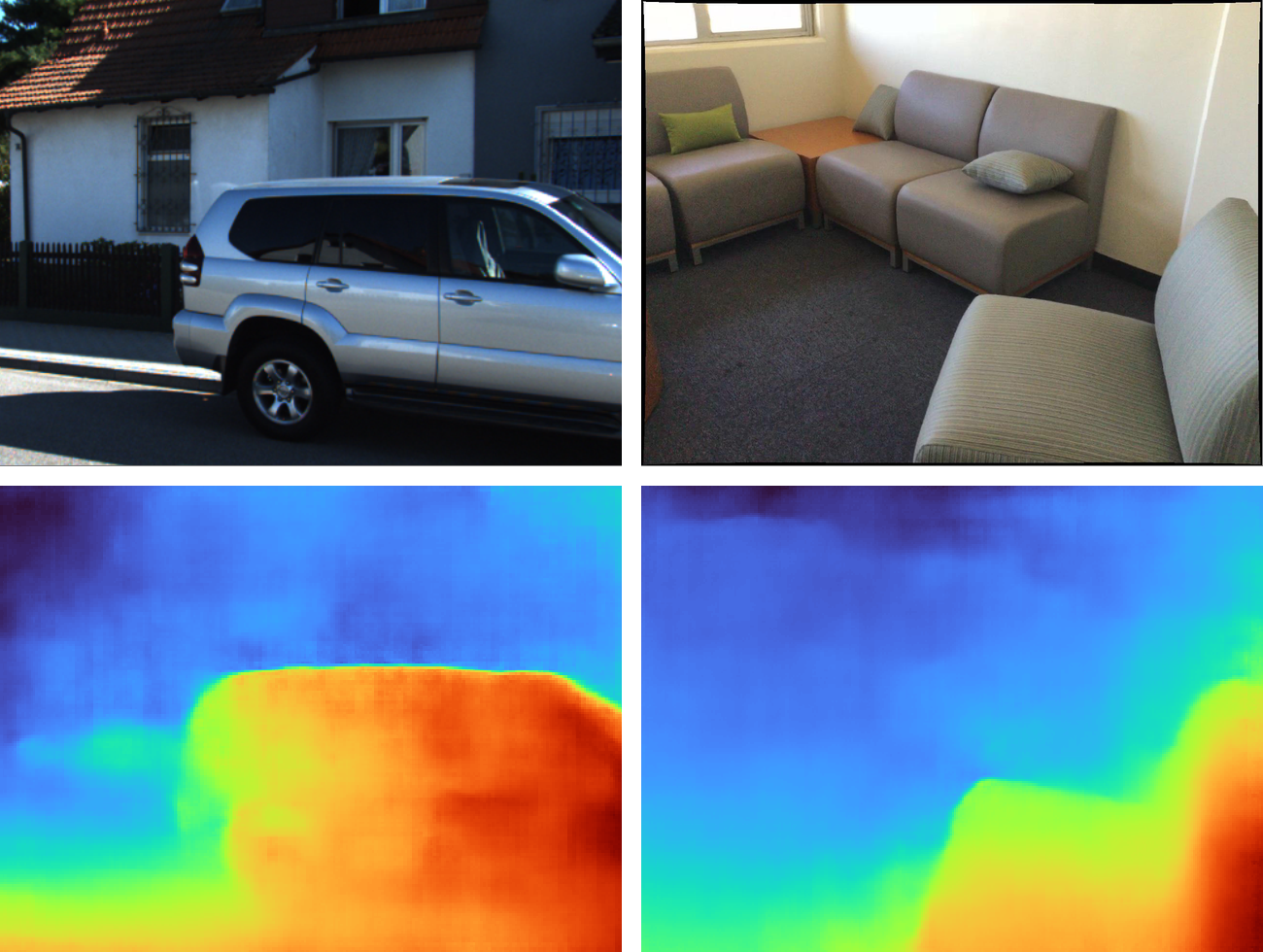}\label{subfig:TrainData-a} }
        \subfloat[]{
        \includegraphics[width=0.49\linewidth]{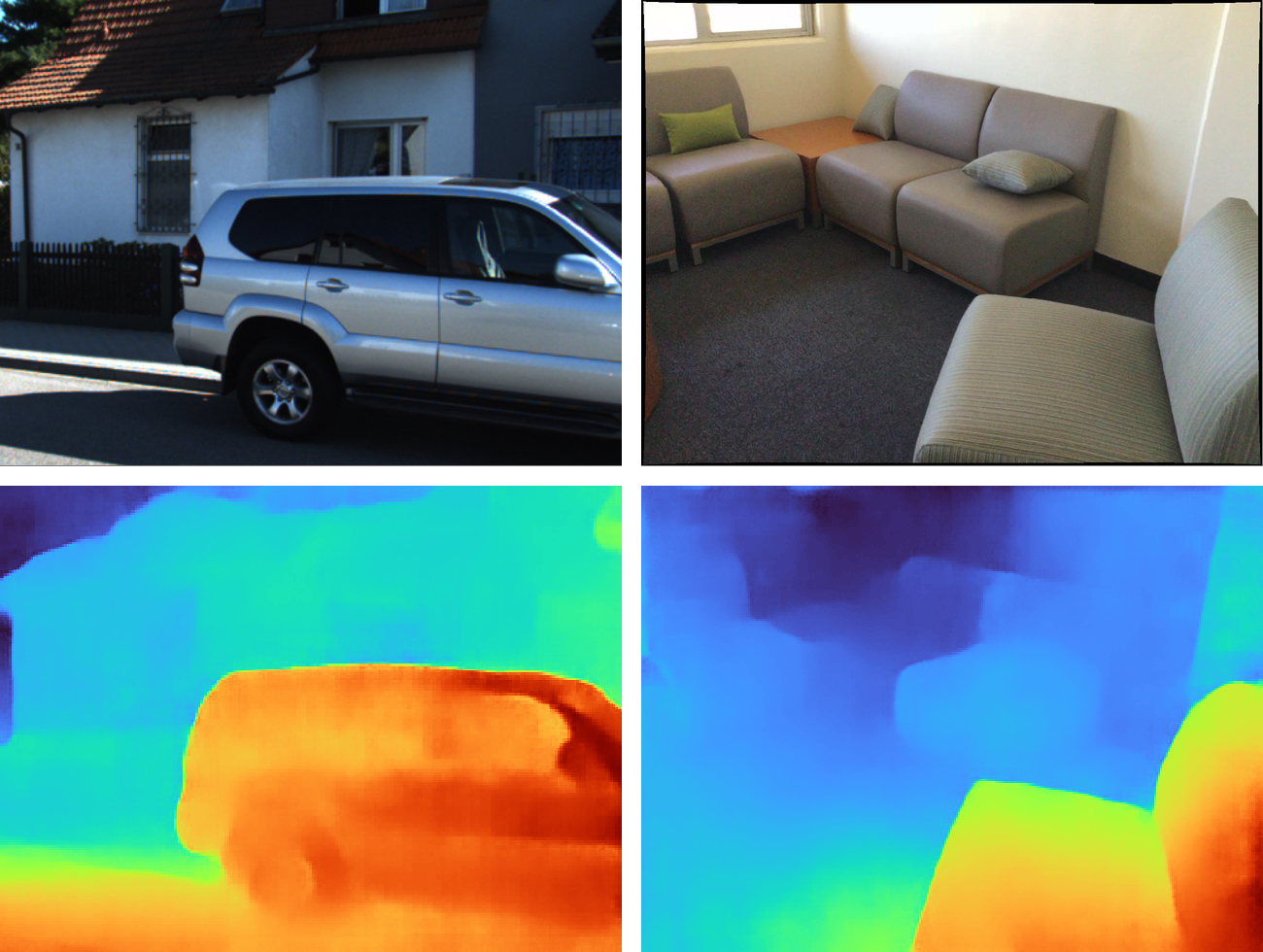}\label{subfig:TrainData-b} }
        \caption{\textbf{Effect of the training dataset}: the first row shows \keyview{} images (\kitti{} and \scannet{}), and the second predicted inverse depth maps. \protect\subref{subfig:TrainData-a} Model trained on \blendedmvs{}.  \protect\subref{subfig:TrainData-b} Model trained jointly on \blendedmvs{}+\staticthings{}.}
        \label{fig:TrainData}
\end{figure}




\subsection{Uncertainty estimation}

\begin{figure}[h!]
        \centering
        \subfloat[]{\includegraphics[width=0.24\linewidth]{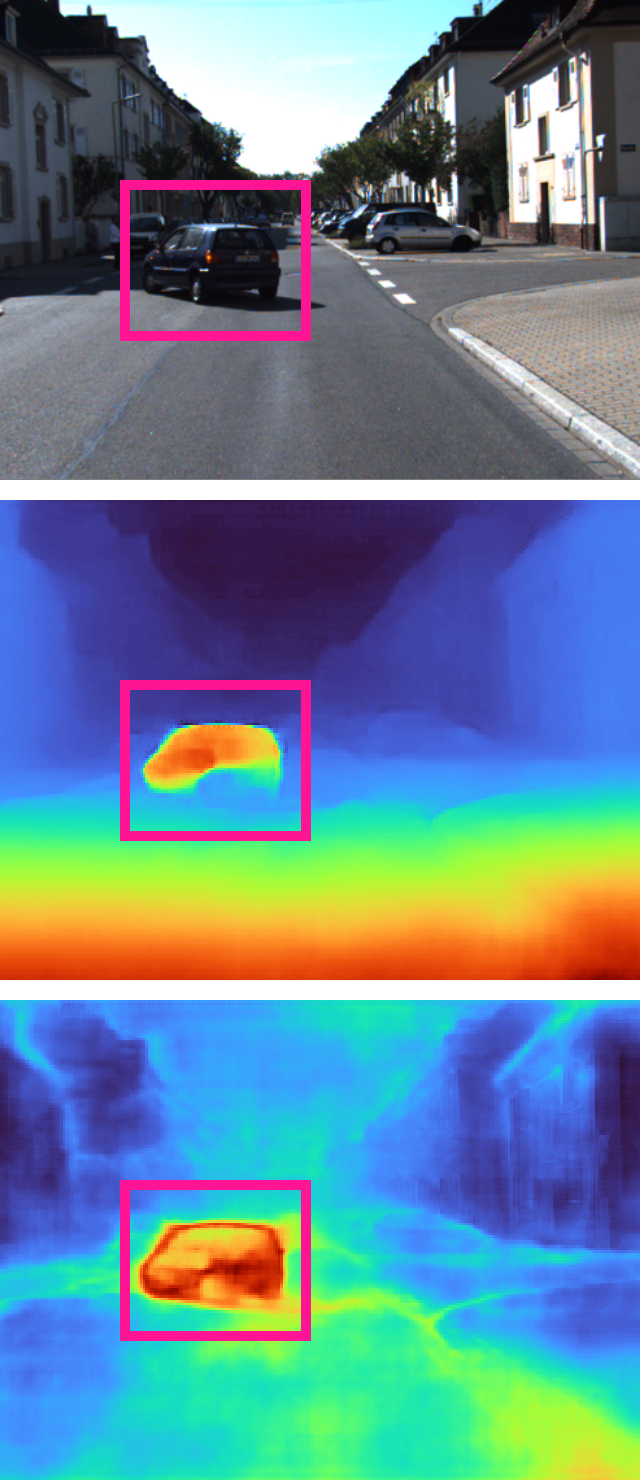}\label{subfig:eval_uncertainty-a} }
        \subfloat[]{
        \includegraphics[width=0.24\linewidth]{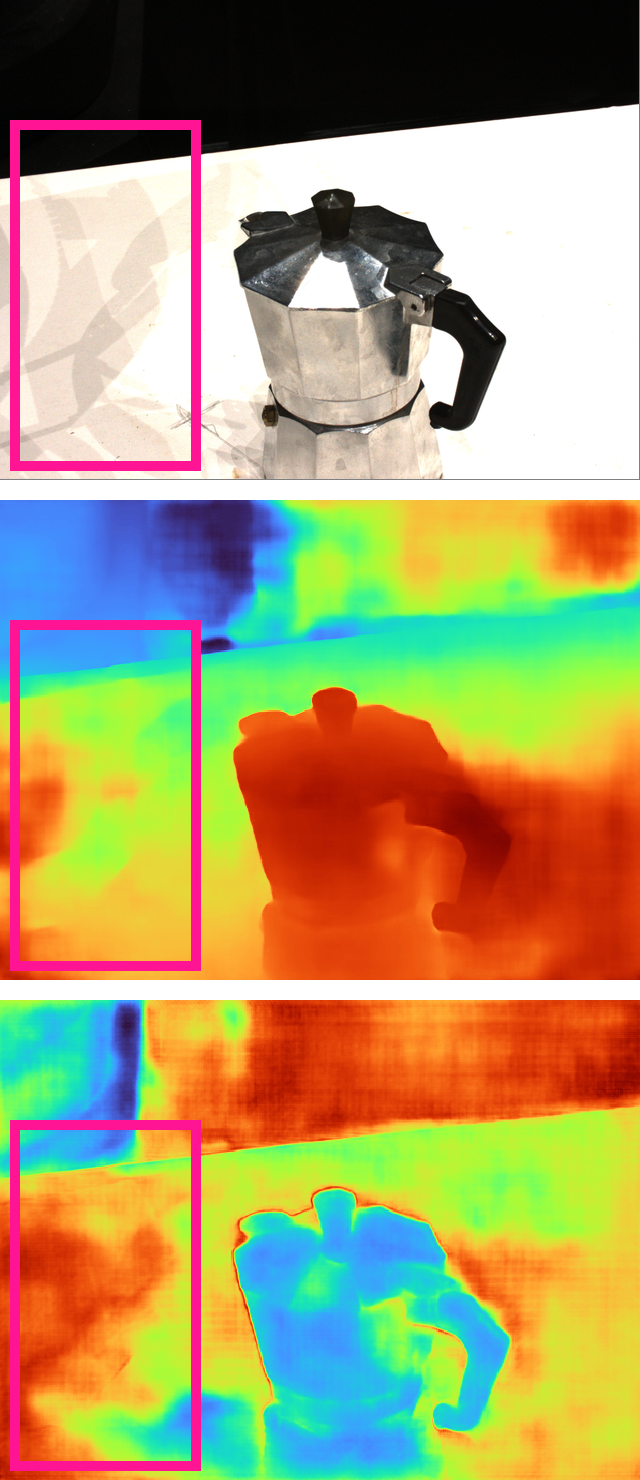}\label{subfig:eval_uncertainty-b} }
        \subfloat[]{
        \includegraphics[width=0.24\linewidth]{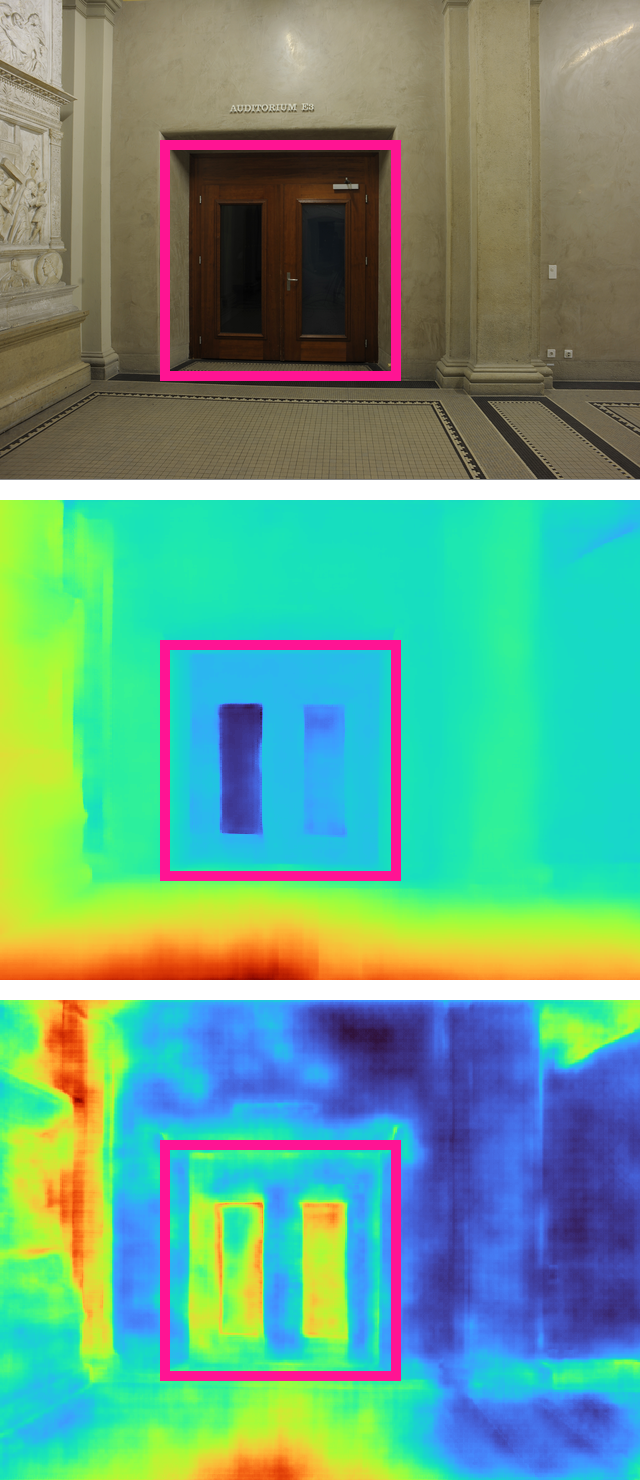}\label{subfig:eval_uncertainty-c} }
        \subfloat[]{
        \includegraphics[width=0.24\linewidth]{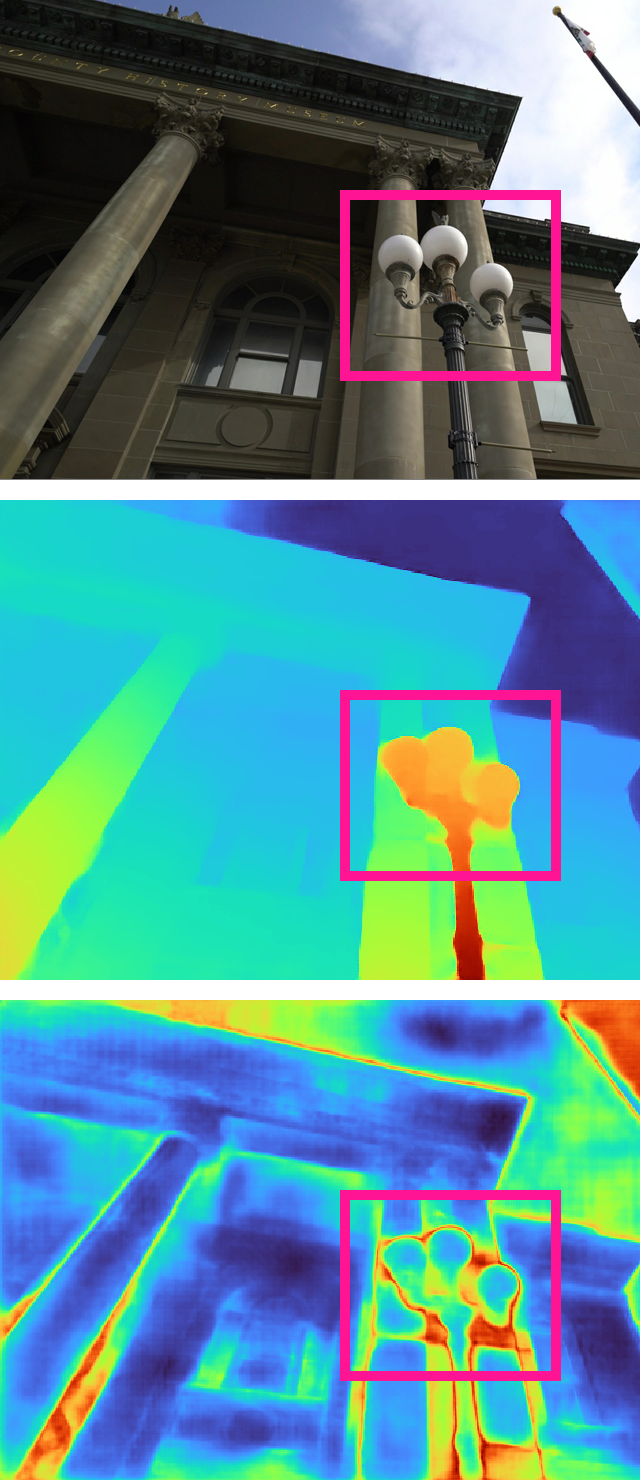}\label{subfig:eval_uncertainty-d}}
        \caption{\textbf{Uncertainty estimation}: the first row shows \keyview{} images, the second predicted inverse depth maps, and the third predicted uncertainty maps (red is uncertain). The model outputs high uncertainties for problematic cases \eg \protect\subref{subfig:eval_uncertainty-a} moving objects,  \protect\subref{subfig:eval_uncertainty-b} textureless regions, \protect\subref{subfig:eval_uncertainty-c} windows, or \protect\subref{subfig:eval_uncertainty-d} fine structures.}
        \label{fig:eval_uncertainty}
\end{figure}

Instead of predicting a point estimate of the inverse depth map, the \baselinename{} model predicts parameters of a Laplace distribution, as in~\cite{Ilg2018} and~\cite{Zhang2020}. For this, an additional output channel is added to the network such that one channel encodes the predicted location parameter and the other the predicted scale parameter. Training is then done by minimizing the negative log likelihood. 
Effects on the depth prediction performance are evaluated in Tab.~\ref{tab:eval_model}d. Predicted uncertainties are evaluated in Tab.~\ref{tab:eval_benchmark_uncertainty} and shown qualitatively in Fig.~\ref{fig:eval_uncertainty}.

\subsection{Multi-view fusion}

We evaluate two strategies for multi-view fusion, namely averaging of cost volumes from multiple \otherviews{}, and weighted averaging with learned weights, \eg as in~\cite{Xu2020}. For the weighted averaging, a small 2D convolutional network with two layers is applied with shared weights to all view-wise cost volumes and outputs pixel-wise weights for each view. We conduct multi-view training with an eraser data augmentation, where regions in \otherviews{} are randomly replaced with the mean color. 
Results for both multi-view fusion strategies are given in Tab.~\ref{tab:eval_model}e. 
The model with learned weights is the \baselinename{} model. 
\section{Conclusion}

We presented the \benchmarkname{} to evaluate the robustness of multi-view depth estimation models on different data domains. The benchmark supports different evaluation settings, \ie different input modalities and optional alignment between predicted and ground truth depths. We found that existing methods have imbalanced performance across domains and cannot be directly applied to arbitrary real-world scenes for estimating depths with their correct scale from given camera poses. We also demonstrated that this can be resolved mostly with existing technology. Together with the benchmark, we provide a robust baseline method that can serve as a basis for future work.

\blfootnote{The research leading to these results is funded by the German Federal Ministry for Economic Affairs and Climate Action within the project ``KI Delta Learning'' (F\"orderkennzeichen 19A19013N). The authors would like to thank the consortium for the successful cooperation. \deeplearningclusterEN. 

Further, we thank \"Ozg\"un Cicek and Christian Zimmermann for their comments on the text and Stefan Teister for keeping our systems running.}

{\small
\bibliographystyle{ieee_fullname}
\bibliography{egbib}
}

\IfFileExists{./01_supplementary.pdf}{%
\newpage


\section*{Supplementary material}

# A Benchmark and a Baseline for Robust Multi-view Depth Estimation
– Appendix –

## A1. Test sets

As specified in the main paper in Sec. 3.1, Test sets, we provide more details on the construction of the test sets for the Robust MVD Benchmark in the following.

**KITTI [17]** The KITTI test set is based on the commonly used Eigen test split [3], which contains 697 samples. More specifically, evaluation is done only on samples of the Eigen split where dense ground truth depth from Uhrig *et al.* [17] is available (652 samples), and where ground truth poses from the KITTI odometry benchmark are available (95 samples). For each sample, sequences with 10 source views before and 10 source views after the keyview are used. This additional restriction leads to the final KITTI test set with 93 samples.

**ETH3D [14]** On ETH3D, the test set is based on the original training split of the high-resolution multi-view benchmark with a total of 13 sequences and 454 views. For the test set, 8 views of each sequence are used as keyviews, resulting in a total of 104 samples. Each sample contains 10 source views, using the view selection provided by https://github.com/FangjinhuaWang/PatchmatchNet [19].

**ScanNet [2]** The ScanNet test set is based on the split from Teed and Deng [16], which in turn extends the split by Tang and Tan [15]. The split consists of 2000 samples taken from 90 of the 1513 totally available sequences. Each sample comprises a keyview and 7 source views, with 3 views before and 4 views after the keyview. For the benchmark test set, every 10th sample of the original split is used, resulting in a total of 200 samples. Images are resized to a resolution of 640x480px to match ground truth depth maps.

**DTU [7, 1]** On DTU, the test set is based on the evaluation split used in MVSNet [23]. The split comprises 22 scans, where each scan has 49 frames. For the benchmark test set, 5 views of each scan are used as keyviews, resulting in a total of 110 samples. Each sample contains 10 source views, using the view selection and depth maps provided by https://github.com/YoYo000/MVSNet [23].

**Tanks and Temples (T&T) [8]** The Tanks and Temples test is based on the scenes "Barn", "Courthouse", "Church", and "Ignatius" from the original training split. Other scenes from the training split were not used, as they were not available for download at the time of writing. To construct the test set, we start with the provided image sets of the respective scenes. We reconstruct camera poses and intrinsics by running COLMAP [13, 12] on the images. The reconstructed camera poses are aligned with the ground truth using the evaluation script from Tanks and Temples (https://github.com/isl-org/TanksAndTemples/tree/master/python_toolbox/evaluation). To obtain ground truth depth maps for each image, we use the provided ground truth pointclouds for each scene and project the points into images. After outlier filtering, we obtain ground truth depth maps, as shown in Fig. A1. To speed up evaluation, only few images per scene are used as keyviews for the test set (Barn: 18; Church: 25; Couthouse: 13; Ignatius: 13), resulting in a total of 69 samples. For each keyview, 10 source views are used, which are selected with the view selection script from https://github.com/FangjinhuaWang/PatchmatchNet/blob/main/colmap_input.py [19].

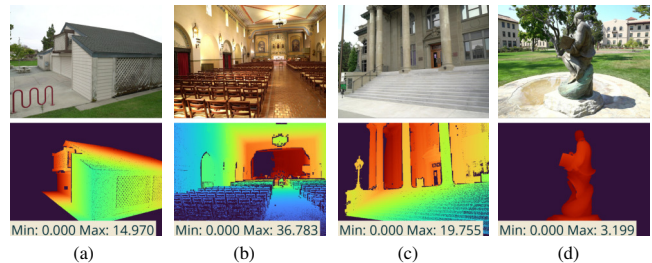

Figure A1. **Tanks and Temples ground truth depth**: the first row shows keyview images, the second ground truth depth maps that were reconstructed from the provided pointclouds. (a) Barn scene. (b) Church scene. (c) Courthouse scene. (d) Ignatius scene.

## A2. Uncertainty estimation metrics

As specified in the main paper in Sec. 3.1, Uncertainty estimation metrics, we provide further details on the uncertainty estimation metrics and evaluation results in the following. In the main paper, we show Sparsification Error Curves for all evaluated models. Additionally, here, we explain and provide Sparsification Curves for all evaluated models, which should help in interpreting the Sparsification

Error Curves.

**Sparsification Curves** Sparsification Curves indicate the error reduction on a metric (here: the Absolute Relative Error) when excluding a certain fraction of pixels from the metric computation based on some ranking. This ranking can be based on estimated uncertainties or on actual errors relative to the ground truth. When using estimated uncertainties for the ranking, a monotonically decreasing Sparsification Curve indicates that uncertainties are aligned with actual errors. The ranking based on actual errors to the ground truth is referred to as Oracle Sparsification Curve and serves as a lower bound, as it represents a hypothetical optimal alignment of uncertainties and actual errors. Hence, smaller differences between both Sparsification Curves indicate better uncertainty estimates. The Sparsification Error Curves from the main paper are the curves that one obtains when taking the difference between both Sparsification Curves.

Both Sparsification Curves (Oracle and uncertainty-based) for different models are shown in Fig. A2 and A3. The difference between both curves are exactly the Sparsification Error Curves of the respective models that are shown in Fig. 3 of the main paper.

Finally, as a side-note, Sparsification can be measured on each sample individually, or on the whole test set. In this work, we compute Sparsification Curves on each sample of a test set individually and then compute the average across all samples in the test set. Following this, we compute the average across all test sets. In Fig. A2 and A3, we plot the average across all test sets in a bold color and indicate the Sparsification Curves on individual test sets in light colors. Additionally, we indicate the standard deviation across test sets as shaded areas in the plots.

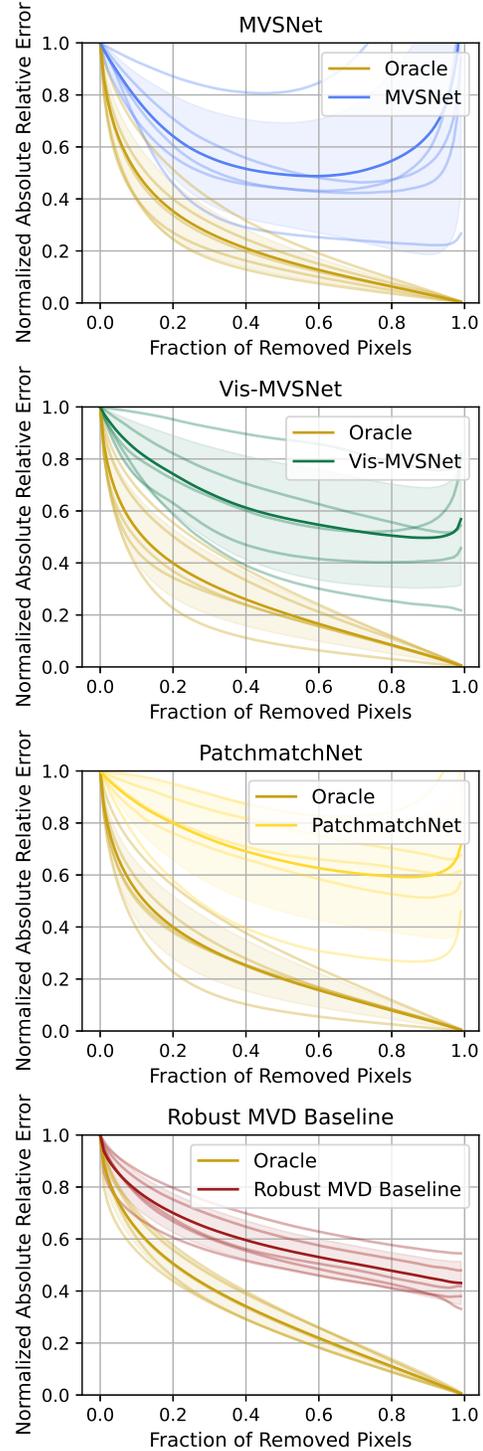

Figure A2. **Sparsification Curves for the Robust MVD Baseline model, MVSNet, Vis-MVSNet and PatchmatchNet.** The difference between both curves is exactly the Sparsification Error Curve from Fig. 3 in the main paper.

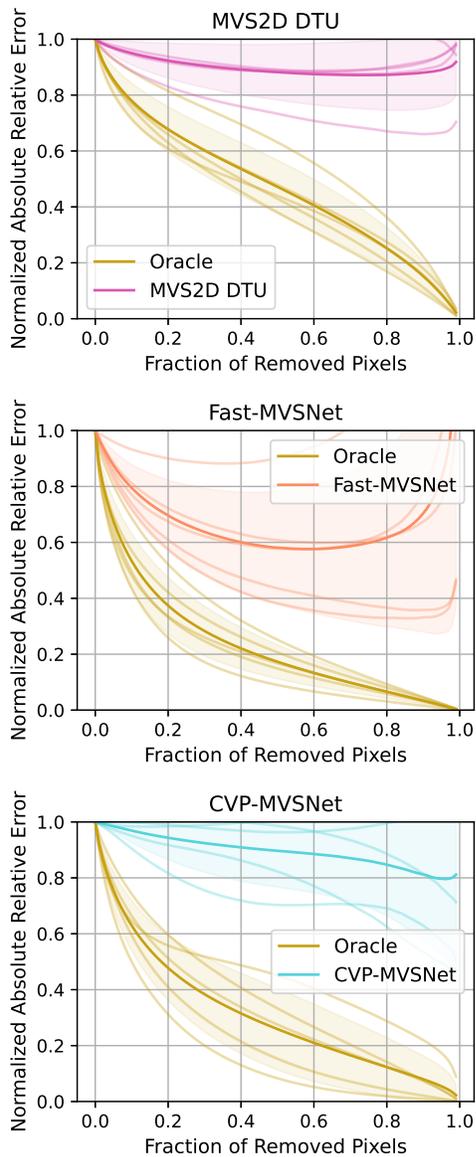

Figure A3. **Sparsification Curves for Fast-MVSNet, CVP-MVSNet and MVS2D.** The difference between both curves is exactly the Sparsification Error Curve from Fig. 3 in the main paper.

## A3. Model runtimes

As specified in Sec. 3.2, Performances depending on source views of the main paper, we plot model runtimes for different numbers of source views in Fig. A4 and A5.

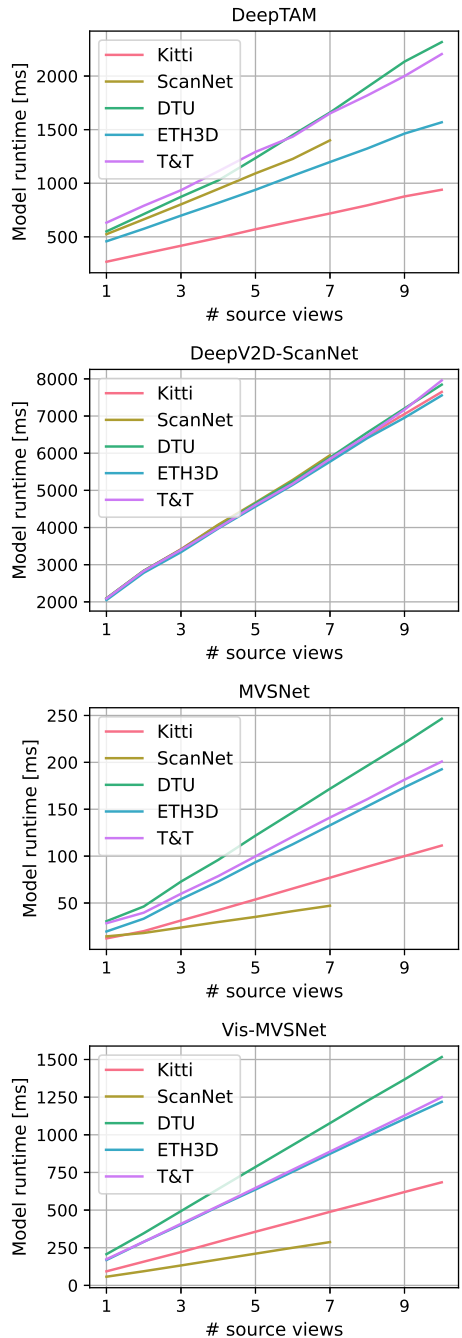

Figure A4. **Model runtimes for different numbers of source views** for DeepTAM, DeepV2D, MVSNet and Vis-MVSNet on all test sets. Different runtimes on the test sets are due to different input resolutions.

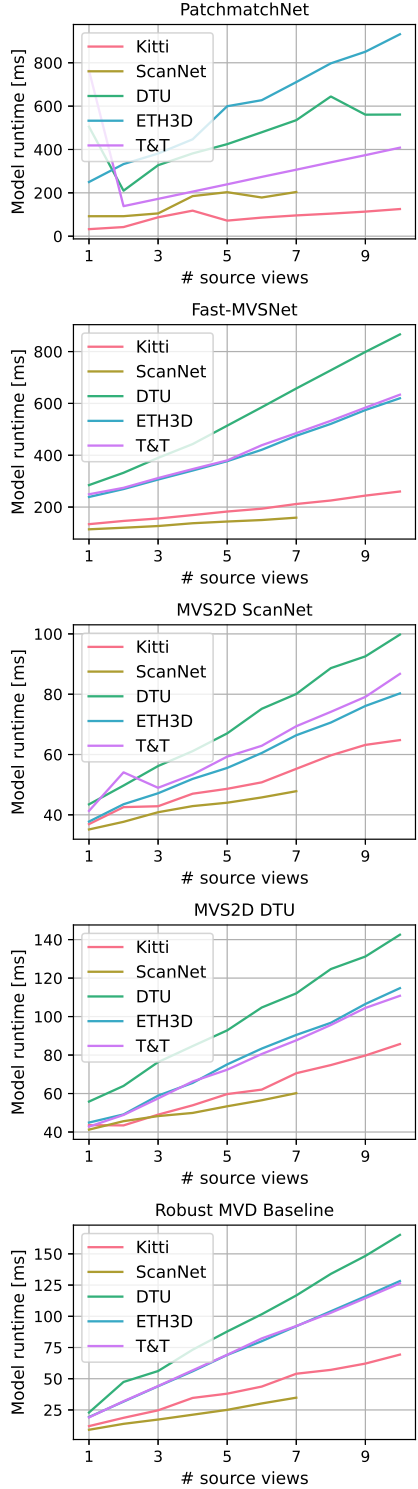

Figure A5. **Model runtimes for different numbers of source views** for PatchmatchNet, Fast-MVSNet, MVS2D and the Robust MVD Baseline model on all test sets. Different runtimes on the test sets are due to different input resolutions.

| Operation | Kernel | Stride | Ch I/O | InpRes | OutRes | Input | Output |
|---|---|---|---|---|---|---|---|
| **(1) Siamese encoder** | | | | | | | |
| for each view $i = 0, ..., k$: | | | | | | | |
| convolution | 7×7 | 2 | 3/64 | 768×384 | 384×192 | Image $\mathbf{I}_i$ | conv1$_i$ |
| convolution | 5×5 | 2 | 64/128 | 384×192 | 192×96 | conv1$_i$ | conv2$_i$ |
| convolution | 3×3 | 2 | 128/256 | 192×96 | 96×48 | conv2$_i$ | conv3a$_i$ |
| **(2) Plane sweep correlation** | | | | | | | |
| for each source view $i = 1, ..., k$: | | | | | | | |
| correlate view 0 and $i$ | - | - | $256/n = 256$ | 96×48 | 96×48 | conv3a$_0$, conv3a$_i$ | costvolume $\mathbf{C}_i$ |
| **(3) Context encoder** | | | | | | | |
| convolution | 1×1 | 1 | 256/32 | 96×48 | 96×48 | conv3a$_0$ | ctx |
| **(4) Costvolume fusion** | | | | | | | |
| 4.1 for each source view $i = 1, ..., k$: | | | | | | | |
| convolution | 3×3 | 1 | 256/128 | 96×48 | 96×48 | $\mathbf{C}_i$ | convfl$_i$ |
| convolution | 3×3 | 1 | 128/1 | 96×48 | 96×48 | convfl$_i$ | weight $\mathbf{w}_i$ |
| 4.2 fuse to costvolume $\mathbf{C}$: $\mathbf{C} = (\sum \exp(\mathbf{w}_i) \cdot \mathbf{C}_i)/(\sum \exp(\mathbf{w}_i))$ | - | - | 256/256 | 96×48 | 96×48 | $\mathbf{C}_i, \mathbf{w}_i$ | $\mathbf{C}$ |
| **(5) Costvolume Decoder** | | | | | | | |
| convolution | 3×3 | 1 | 288/256 | 96×48 | 96×48 | $\mathbf{C}$+ctx | conv3b |
| convolution | 3×3 | 2 | 256/512 | 96×48 | 48×24 | conv3b | conv4a |
| convolution | 3×3 | 1 | 512/512 | 48×24 | 48×24 | conv4a | conv4b |
| convolution | 3×3 | 2 | 512/512 | 48×24 | 24×12 | conv4b | conv5a |
| convolution | 3×3 | 1 | 512/512 | 24×12 | 24×12 | conv5a | conv5b |
| convolution | 3×3 | 2 | 512/1024 | 24×12 | 12×6 | conv5b | conv6a |
| convolution | 3×3 | 1 | 1024/1024 | 12×6 | 12×6 | conv6a | conv6b |
| convolution | 3×3 | 1 | 1024/2 | 12×6 | 12×6 | conv6b | pred6 |
| transposed convolution | 4×4 | 2 | 1024/512 | 12×6 | 24×12 | conv6b | upconv5 |
| convolution | 3×3 | 1 | 1025/512 | 24×12 | 24×12 | upconv5+pr6+conv5b | iconv5 |
| convolution | 3×3 | 1 | 512/2 | 24×12 | 24×12 | iconv5 | pred5 |
| transposed convolution | 4×4 | 2 | 512/256 | 24×12 | 48×24 | iconv5 | upconv4 |
| convolution | 3×3 | 1 | 769/256 | 48×24 | 48×24 | upconv4+pr5+conv4b | iconv4 |
| convolution | 3×3 | 1 | 256/2 | 48×24 | 48×24 | iconv4 | pred4 |
| transposed convolution | 4×4 | 2 | 256/128 | 48×24 | 96×48 | iconv4 | upconv3 |
| convolution | 3×3 | 1 | 385/128 | 96×48 | 96×48 | upconv3+pr4+conv3b | iconv3 |
| convolution | 3×3 | 1 | 128/2 | 96×48 | 96×48 | iconv3 | pred3 |
| transposed convolution | 4×4 | 2 | 128/64 | 96×48 | 192×96 | iconv3 | upconv2 |
| convolution | 3×3 | 1 | 193/64 | 192×96 | 192×96 | upconv2+pr3+conv2$_0$ | iconv2 |
| convolution | 3×3 | 1 | 64/2 | 192×96 | 192×96 | iconv2 | pred2 |
| transposed convolution | 4×4 | 2 | 64/32 | 192×96 | 384×192 | iconv2 | upconv1 |
| convolution | 3×3 | 1 | 97/32 | 384×192 | 384×192 | upconv1+pr2+conv1$_0$ | iconv1 |
| convolution | 3×3 | 1 | 32/2 | 384×192 | 384×192 | iconv1 | pred1 |

Table A1. **Architecture of the Robust MVD Baseline model**. The model is based on a DispNet architecture. The notation is the same as in the main paper, *i.e.* the model takes views $V = (V_0, \cdots, V_k)$ as input where $V_0$ is the keyview with an image $\mathbf{I}_0$ and $V_{1,..,k}$ are source views with images $\mathbf{I}_{i,..,k}$. $n$ is the number of candidate inverse depth values in the plane sweep correlation layer (here: 256). The resolutions are indicated for training settings and differ for test inputs. The table is adapted from [10].

## A4. Model architecture

As specified in Sec. 4.1, Model architecture of the main paper, we provide further details on the model architecture in the following.

The architecture is based on a standard DispNetC architecture [10] with the following adaptations: (1) the correlation layer is replaced by a plane sweep correlation layer, (2) costvolumes from multiple views are fused via weighted averaging with learned weights, and (3) all outputs have two channels instead of one, with the additional channel being the scale parameter of the predicted Laplace distribution. A summary of the architecture is provided in Tab. A1.

Instead of the original DispNet correlation layer, the model correlates in a plane sweep stereo fashion, which has been shown to work well in other multi-view depth architectures [16, 27, 23, 9, 5]. For each feature $\mathbf{f}_0$ at a location $^0\mathbf{x}$ in the keyview $V_0$, points $^i\mathbf{x}$ in a source view $V_i$ are sampled by reprojecting $^0\mathbf{x}$ to the view $V_i$ image plane for different candidate inverse depth values $d$ as follows [4]:

$$^i\tilde{\mathbf{x}}(d) = \mathbf{K}_i \cdot {}^i_0\mathbf{R} \cdot \mathbf{K}_0^{-1} \cdot {}^0\tilde{\mathbf{x}} + \mathbf{K}_i \cdot {}^i_0\mathbf{t} \cdot d, \quad (1)$$

with $^0\tilde{\mathbf{x}}$ and $^i\tilde{\mathbf{x}}$ denoting $^0\mathbf{x}$ and $^i\mathbf{x}$ in homogeneous coordinates. We use $n = 256$ candidate inverse depth values $d$ that are spaced equidistantly in the range $d_{min} = 0.001 \, \text{m}^{-1}$ to $d_{max} = 2.5 \, \text{m}^{-1}$. Given these sampling points, features

| Approach | GT Poses | GT Range | Align | StaticThings3D | | | | BlendedMVS | | | |
|---|---|---|---|---|---|---|---|---|---|---|---|
| | | | | rel ↓ | τ ↑ | AUSE ↓ | time [s] | rel ↓ | τ ↑ | AUSE ↓ | time [s] |
| **a) Classical:** | | | | | | | | | | | |
| COLMAP [13, 12] | ✓ | ✗ | ✗ | 5.8 | 79.5 | - | ≈ 2 min | 1.4 | 97.2 | - | ≈ 2 min |
| COLMAP Dense [13, 12] | ✓ | ✗ | ✗ | 16.6 | 65.7 | - | ≈ 2 min | 4.2 | 92.3 | - | ≈ 2 min |
| **b)** | | | | | | | | | | | |
| DeMoN [18] | ✗ | ✗ | $\|\mathbf{t}\|$ | 36.4 | 6.3 | - | 0.07 | 34.4 | 14.5 | - | 0.08 |
| DeepV2D ScanNet [16] | ✗ | ✗ | med | 23.1 | 15.4 | - | 2.73 | - | - | - | - |
| DeepV2D KITTI [16] | ✗ | ✗ | med | - | - | - | - | - | - | - | - |
| **c)** | | | | | | | | | | | |
| MVSNet [23] | ✓ | ✓ | ✗ | 38.9 | 39.0 | 0.45 | 0.06 | 5.0 | 80.7 | 0.30 | 0.03 |
| MVSNet Inv. Depth [23] | ✓ | ✓ | ✗ | 26.5 | 43.3 | 0.59 | 0.12 | 46.1 | 14.8 | 0.38 | 0.07 |
| Vis-MVSNet [26] | ✓ | ✓ | ✗ | 10.4 | 59.3 | 0.42 | 0.47 | 2.5 | 92.6 | 0.20 | 0.47 |
| CVP-MVSNet [21] | ✓ | ✓ | ✗ | - | - | - | - | - | - | - | - |
| PatchmatchNet [19] | ✓ | ✓ | ✗ | 7.6 | 58.1 | 0.58 | 0.07 | 22.1 | 27.9 | 0.43 | 0.05 |
| Fast-MVSNet [25] | ✓ | ✓ | ✗ | 24.9 | 42.7 | 0.52 | 0.26 | 13.0 | 59.2 | 0.45 | 0.11 |
| MVS2D DTU [22] | ✓ | ✓ | ✗ | 88.3 | 2.6 | 0.52 | 0.04 | 38.9 | 7.3 | 0.54 | 0.04 |
| MVS2D ScanNet [22] | ✓ | ✓ | ✗ | 39.9 | 2.7 | - | 0.04 | 64.3 | 1.6 | - | 0.04 |
| **d) Absolute scale:** | | | | | | | | | | | |
| DeMoN [18] | ✓ | ✗ | ✗ | 36.3 | 6.2 | - | 0.08 | 44.2 | 7.6 | - | 0.08 |
| DeepTAM [27] | ✓ | ✗ | ✗ | - | - | - | - | - | - | - | - |
| DeepV2D KITTI [16] | ✓ | ✗ | ✗ | - | - | - | - | - | - | - | - |
| DeepV2D ScanNet [16] | ✓ | ✗ | ✗ | 86.2 | 1.8 | - | 1.92 | - | - | - | - |
| MVSNet [23] | ✓ | ✗ | ✗ | 26.1 | 42.2 | 0.48 | 0.12 | - | - | - | - |
| MVSNet Inv. Depth [23] | ✓ | ✗ | ✗ | 53.4 | 4.7 | - | 0.14 | 29.1 | 48.7 | - | 0.07 |
| CVP-MVSNet [21] | ✓ | ✗ | ✗ | - | - | - | - | - | - | - | - |
| Vis-MVSNet [26] | ✓ | ✗ | ✗ | 9.6 | 59.2 | 0.39 | 0.47 | - | - | - | - |
| PatchmatchNet [19] | ✓ | ✗ | ✗ | 42.5 | 2.6 | 0.36 | 0.04 | 44.7 | 8.4 | 0.38 | 0.06 |
| Fast-MVSNet [25] | ✓ | ✗ | ✗ | 26.1 | 35.9 | - | 0.21 | 174.5 | 15.8 | - | 0.19 |
| MVS2D ScanNet [22] | ✓ | ✗ | ✗ | 92.4 | 0.0 | - | 0.04 | 29.9 | 52.3 | - | 0.05 |
| MVS2D DTU [22] | ✓ | ✗ | ✗ | 97.2 | 0.0 | 0.02 | 0.04 | 45.6 | 21.1 | 0.19 | 0.04 |
| Robust MVD Baseline | ✓ | ✗ | ✗ | 6.3 | 54.7 | 0.19 | 0.04 | 3.4 | 81.4 | 0.20 | 0.04 |

Table A2. **Quantitative results on StaticThings3D and BlendedMVS test sets.** Results are reported for the Absolute Relative Error (rel), the Inlier Ratio ($\tau$) with a threshold of 1.25, the Area Under Sparsification Error Curve (AUSE) and the model runtime in seconds. Results are for the quasi-optimal selection of source views. COLMAP results are not directly comparable due to a lower prediction density.

$\mathbf{f}_i = \mathbf{F}_i(^i\mathbf{x})$ are sampled from the view $V_i$ feature map $\mathbf{F}_i$ with bilinear interpolation. The sampled features are compared to the keyview feature with a dot product $\mathbf{f}_0 \cdot \mathbf{f}_i$, resulting in correlation scores for all candidate inverse depth values. This is done for all keyview features, resulting in a costvolume $\mathbf{C}_i$ with pixel-wise correlation scores.

The costvolumes from multiple source views are fused via weighted averaging and mapped to the predicted inverse depth map and uncertainty map by the costvolume decoder network.

## A5. Training details

Training is done with four source views. On StaticThings3D, we randomly sample source views within a range of $\pm 3$ frames around the keyview. On BlendedMVS, we randomly sample from the ten best source views as in [19]. Other than this, we use similar hyperparameters as in the original DispNet, *e.g.* train for 600k iterations (ca. 60 hours on a single 3090 RTX GPU), use the Adam optimizer, start with a learning rate of 1e-4 and decay it by a factor of 0.5 after 300k, 400k and 500k iterations. The network outputs predicted inverse depth maps at different resolutions and is trained with a negative log-likelihood loss for every resolution (see [6]).

## A6. StaticThings3D and BlendedMVS Results

The proposed Robust MVD Baseline model is trained on StaticThings3D and BlendedMVS. In the main paper, we did not evaluate on these datasets, as these datasets are not commonly used for evaluation and as StaticThings3D has little overlap with real-world data. We provide results on test sets from these two datasets in the following.

**StaticThings3D (ST3D) test set** The StaticThings3D test split contains 600 sequences with 10 views each. We define one sample per sequence, using the fifth view as keyview plus 4 source views before and 5 after the keyview, resulting in 600 samples. To speed up evaluation, we use every 10th sample for the test set, resulting in a total of 60 samples.

**BlendedMVS (BMVS) [24] test set** The BlendedMVS test set is based on the original validation split, which contains 7 scenes with a total of 915 frames. For each frame, a selection of 10 source views is provided by the authors. For our test set, we use 10 frames per scene as keyviews, resulting in a total of 70 samples.

**Results** Results are provided in Tab. A2.

## A7. Application to real-world data with known poses

A motivation for the Robust MVD Benchmark were real-world applications where camera poses are known and the objective is to reconstruct depth maps at real-world scale. To illustrate this, we conduct experiments on a small self-captured datasets that we term RealThings. RealThings is captured indoors with a handheld Intel RealSense D435 RGB-D camera and uses April Tags [20] for extrinsic calibration. This can be seen as a typical scenario in robotics, where it is not uncommon to use April Tags.

Qualitative outputs of the Robust MVD Baseline model are shown in Fig. A6. The model was applied with a fixed number of 6 input views and without changing any configurations. This shows that the model can be directly applied for depth estimation in real-world scenarios.

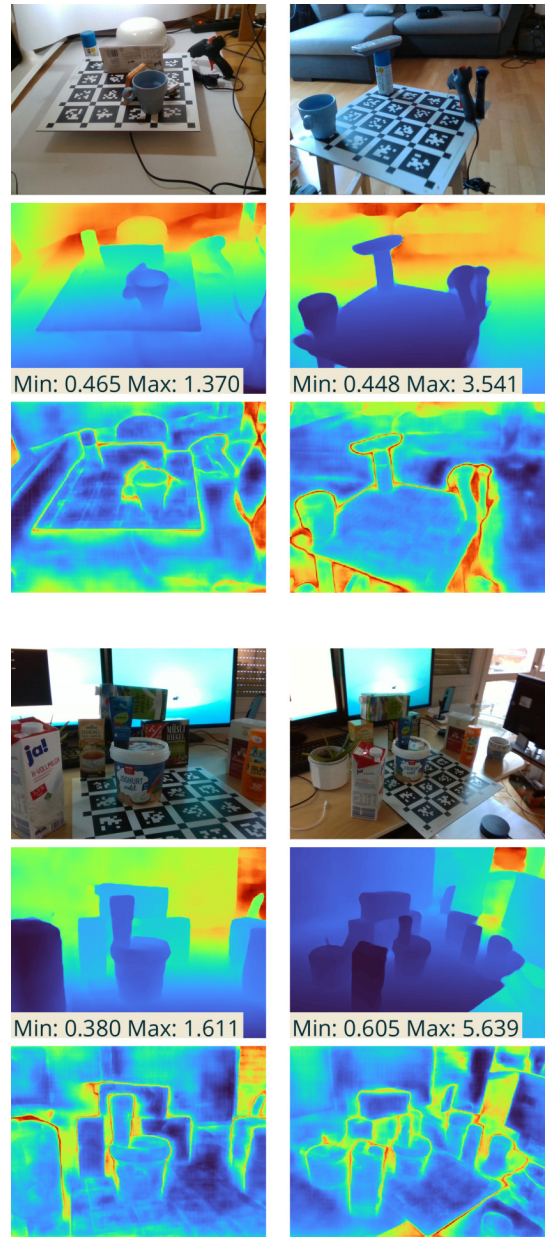

Figure A6. **Qualitative results of the Robust MVD Baseline on samples from the self-captured RealThings dataset.** For each sample, the figure shows three images: the first row shows the keyview image, the second row the predicted depth map, and the third row the predicted uncertainty. This example shows real-world application in a scenario where camera poses are known, *e.g.* from April Tags (tags are visible in the images). In such scenarios, the proposed Robust MVD Baseline model can be applied without any reconfiguration to derive depth maps at real-world scale. Blue colors indicate small values, red colors large values, and the ranges are given in meters.

## A8. Limitations

The proposed Robust MVD Baseline model works well across domains and scales and gives reasonable predictions for most inputs. However, there is still room for improvement regarding overall accuracy of the predictions. Furthermore, the model fails for cases where it is difficult to match between the keyview image and source view images, *e.g.* specular reflections. Usually this results in incorrect depth estimates that are at least indicated by high uncertainties. However, in some cases (probably because incorrect matches due to reflections are consistent across source views), the erroneous depth estimates are not well represented by the estimated uncertainty (*e.g.* the TV screen in Fig. A7). Other obvious failure cases are common problems of multi-view depth estimation, namely small camera motion and dynamically moving objects. These cases could be resolved through stronger single-image priors [11], or by explicit detection and motion estimation for dynamic objects. The proposed model is designed to serve as a baseline on the proposed Robust MVD Benchmark and we look forward to future work that improves upon it. Examples for problematic cases are shown in Fig. A7.

## A9. Qualitative results

In Fig. A8, Fig. A9, and Fig. A10, we show qualitative results of evaluated models. For models from related works, we show qualitatives for the three evaluation settings, namely for depth-from-video, multi-view stereo and absolute scale depth estimation.

In Fig. A11 and Fig. A12, we show additional qualitative results of the proposed Robust MVD Baseline model on samples from different domains and scene scales.

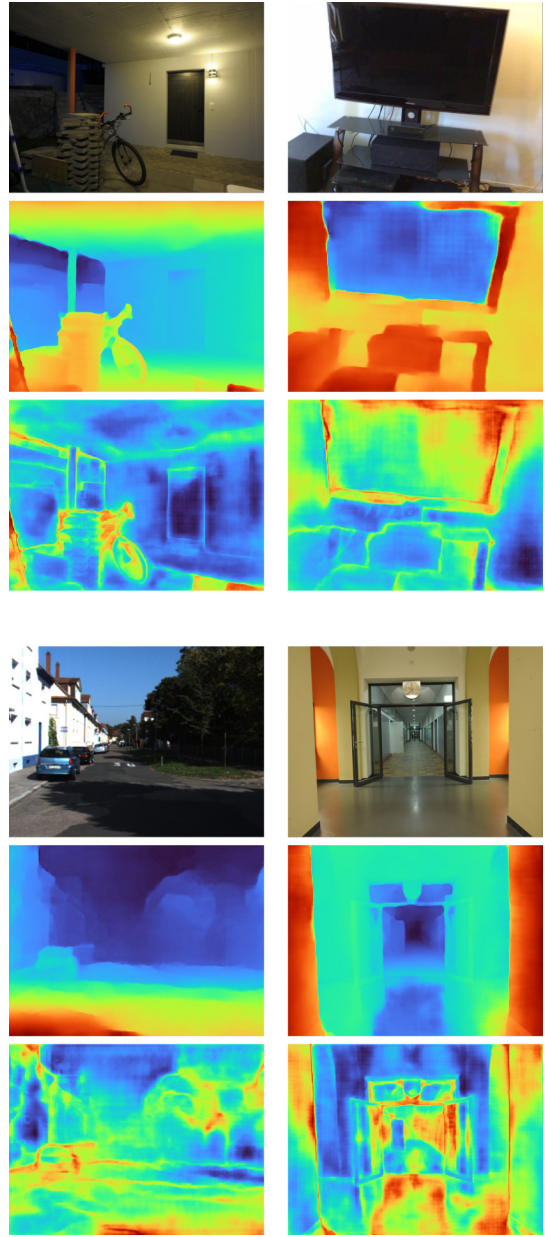

Figure A7. **Limitations of the Robust MVD Baseline.** For each sample, the figure shows three images: the first row shows the keyview image, the second the predicted inverse depth map, and the third the predicted uncertainty. The figure illustrates problematic cases where predictions are inaccurate, *e.g.* for fine structures (bike strokes in the first sample), reflective surfaces (screen in the second sample; floor in the last sample), or strong gradients in the color image (shadow on the road in the third sample).

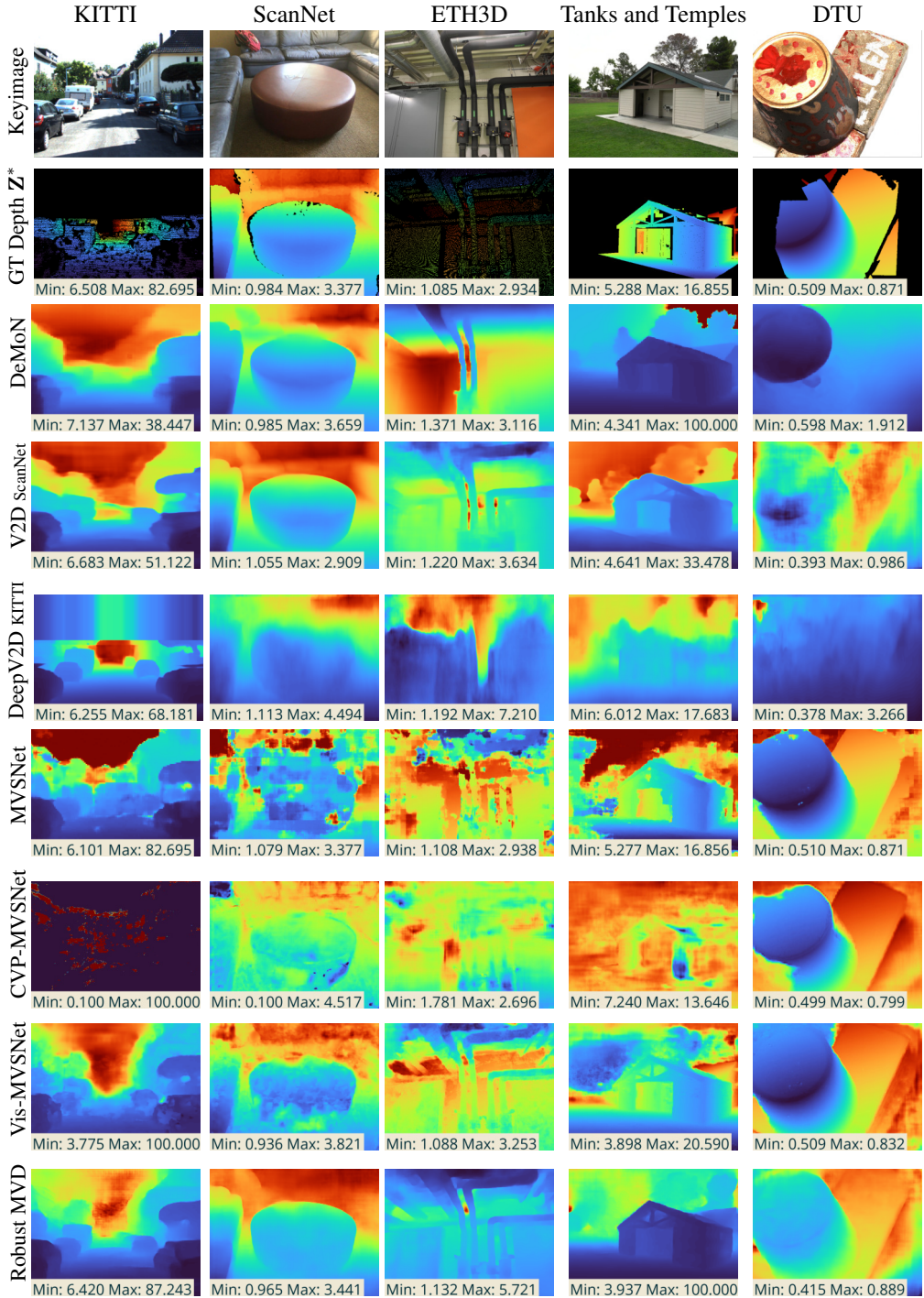

Figure A8. **Predicted depth maps of evaluated models on all test sets.** Models from related works are evaluated up to a relative scale, *i.e.* DeMoN and DeepV2D (V2D) are evaluated in the depth-from-video setting (without ground truth poses, predicted depths are aligned to the ground truth) and MVSNet, CVP-MVSNet and Vis-MVSNET are evaluated in the multi-view stereo setting (ground truth poses and ground truth depth range are provided to the model). The predictions hence correspond to the quantitative results in Tab. 2b and Tab. 2c of the main paper. The Robust MVD model is evaluated in the absolute depth setting (ground truth poses are provided to the model, no alignment between predicted and ground truth depths; Tab. 2d). All models from related works perform significantly better on data from their respective training domain. Blue colors indicate small values, red colors large values, and the ranges are given in meters. Corresponding predicted uncertainties are shown in Fig. A9.

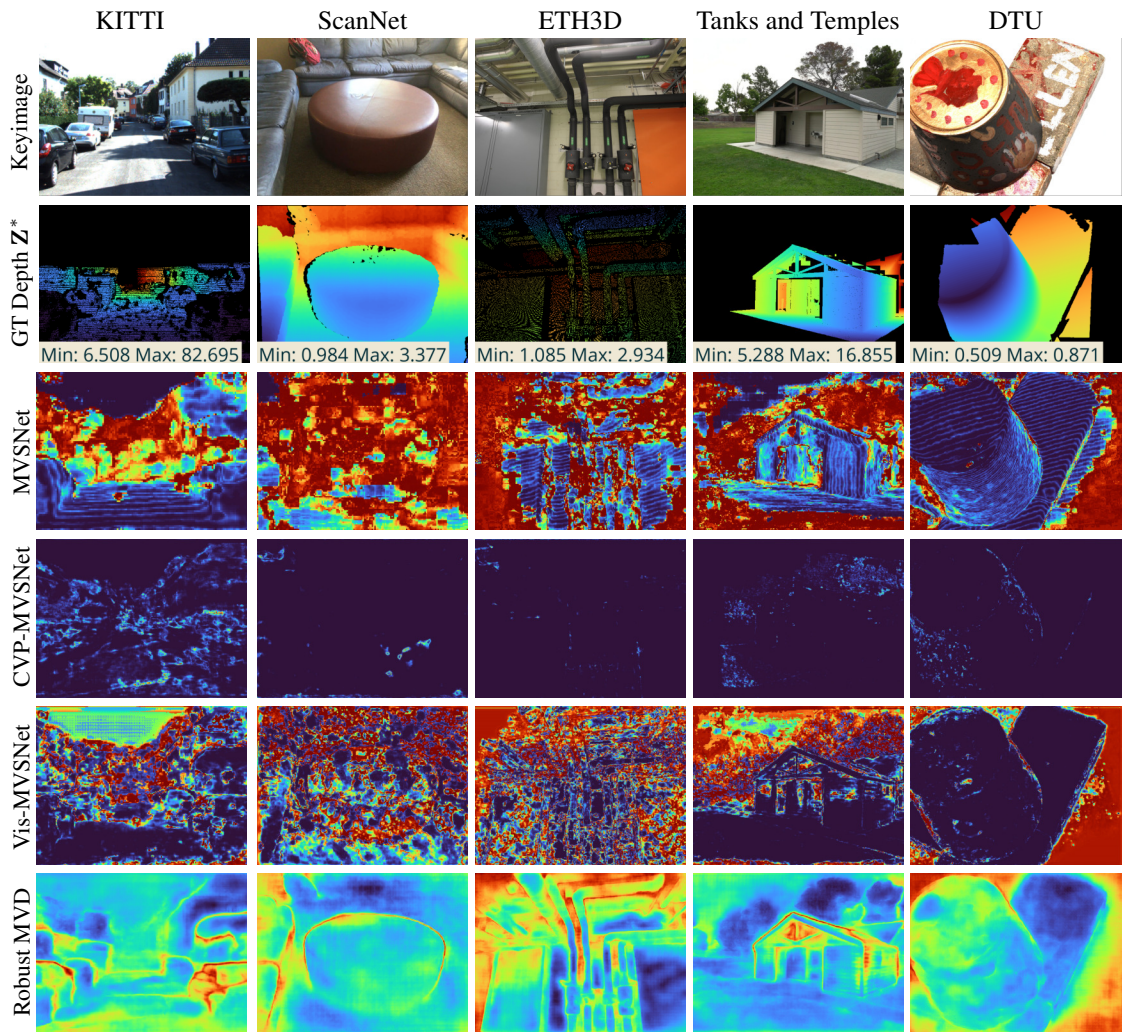

Figure A9. **Predicted uncertainties of evaluated models on all test sets.** Models from related works are evaluated up to a relative scale. The shown uncertainties hence correspond to the depth maps shown in Fig. A8 on the previous page and the quantitative results in Tab. 3 of the main paper. Blue colors indicate low uncertainties, red colors high uncertainties.

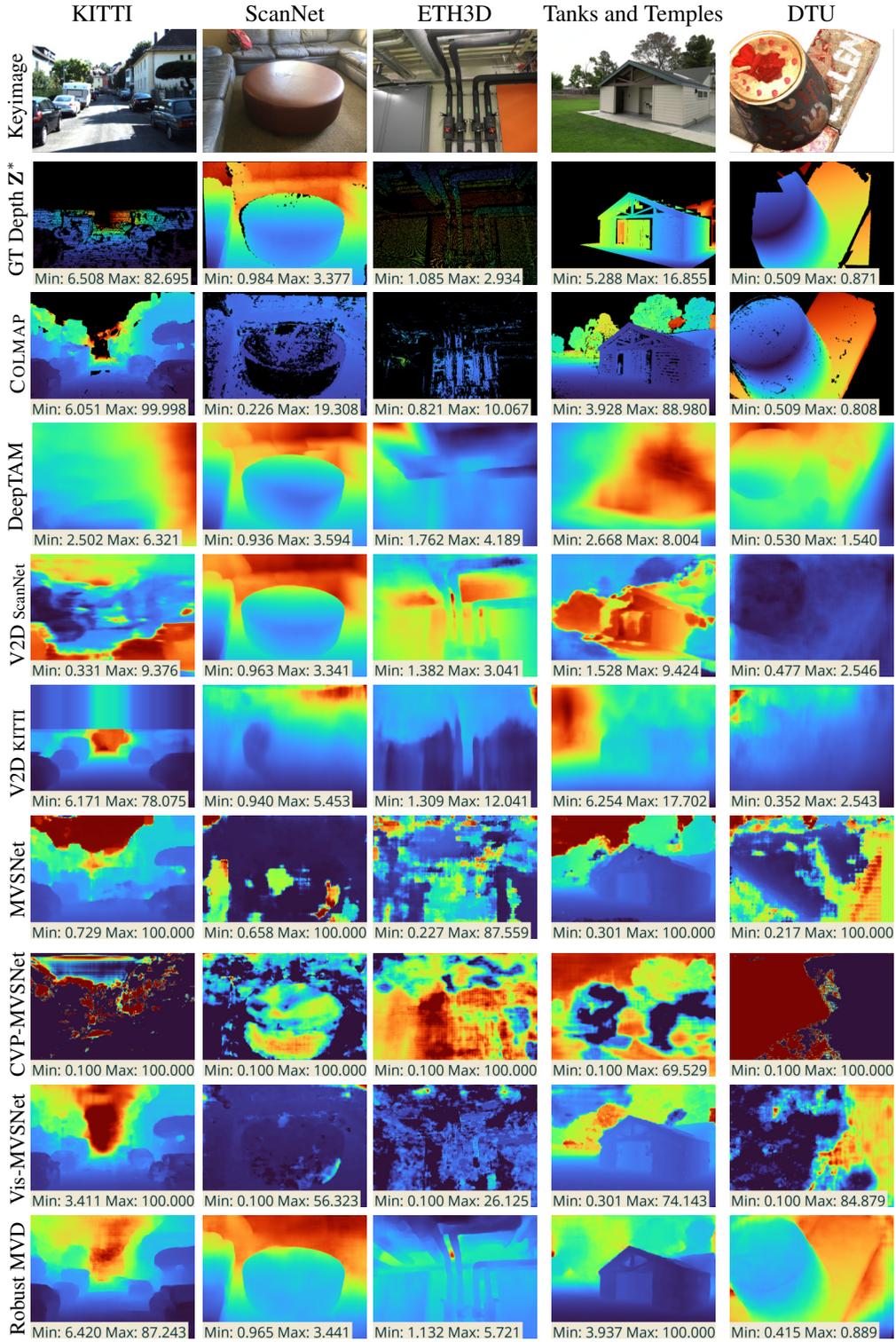

Figure A10. **Predicted depth maps of evaluated models on all test sets in the absolute depth estimation setting.** All models are applied with ground truth poses, without ground truth depth range, and without alignment between predicted and ground truth depths. The predictions shown here hence correspond to the quantitative results in Tab. 2d of the main paper. All models from related works overfit to the costvolume distribution seen during training and perform worse on inputs with different distributions. Blue colors indicate small values, red colors large values, and the ranges are given in meters.

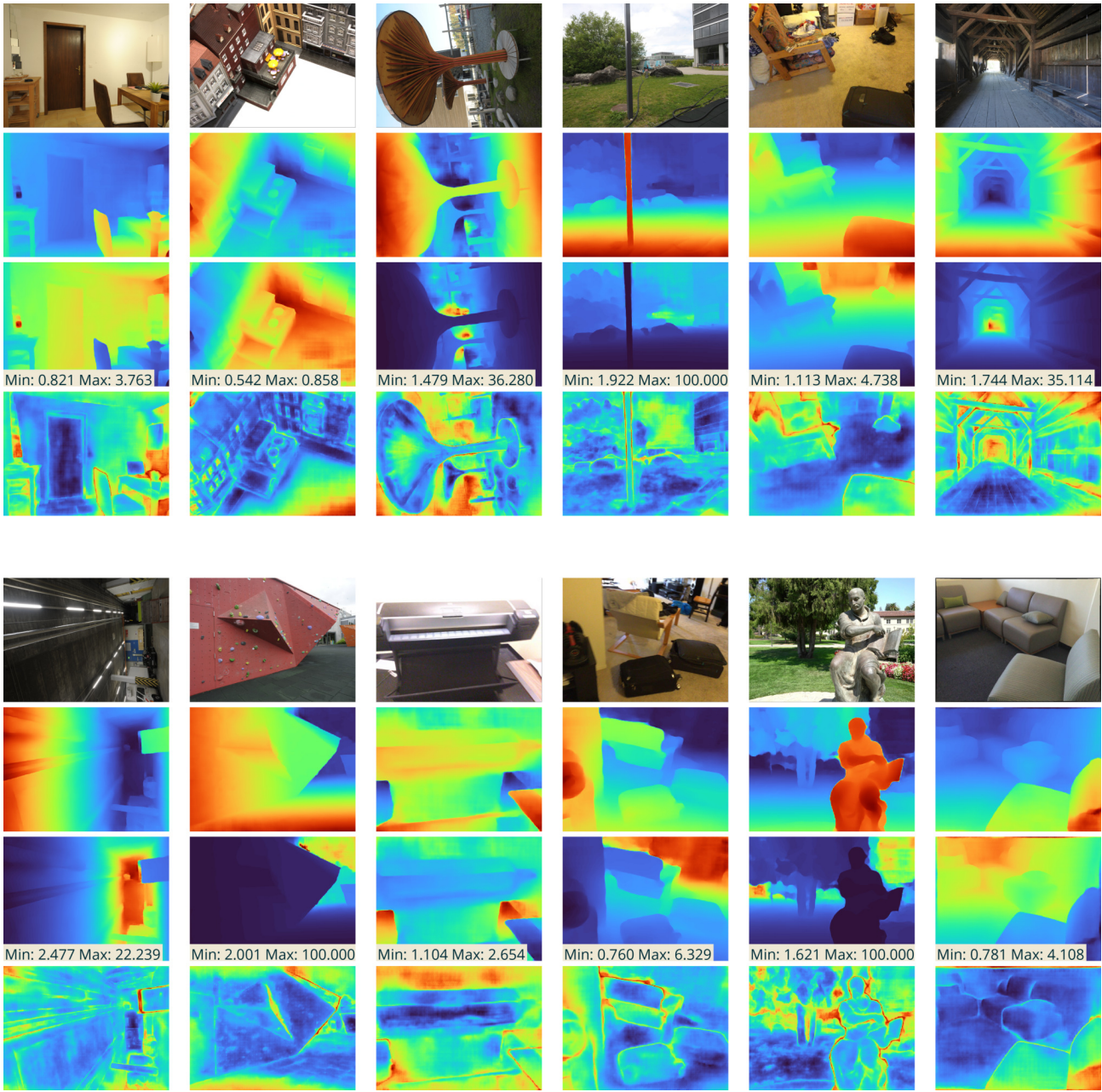

Figure A11. **Qualitative results of the Robust MVD Baseline** on samples from the KITTI, ScanNet, ETH3D, DTU, Tanks and Temples datasets. For each sample, the figure shows four images: the first row shows the keyview image, the second row the predicted inverse depth map, the third row the predicted depth map, and the fourth row the predicted uncertainty. Predicted inverse depth maps and predicted depth maps are equivalent. However, inverse depth maps are better suited for visualization, whereas depth maps allow for an easier interpretation of the scale of predicted values (here indicated in meters).

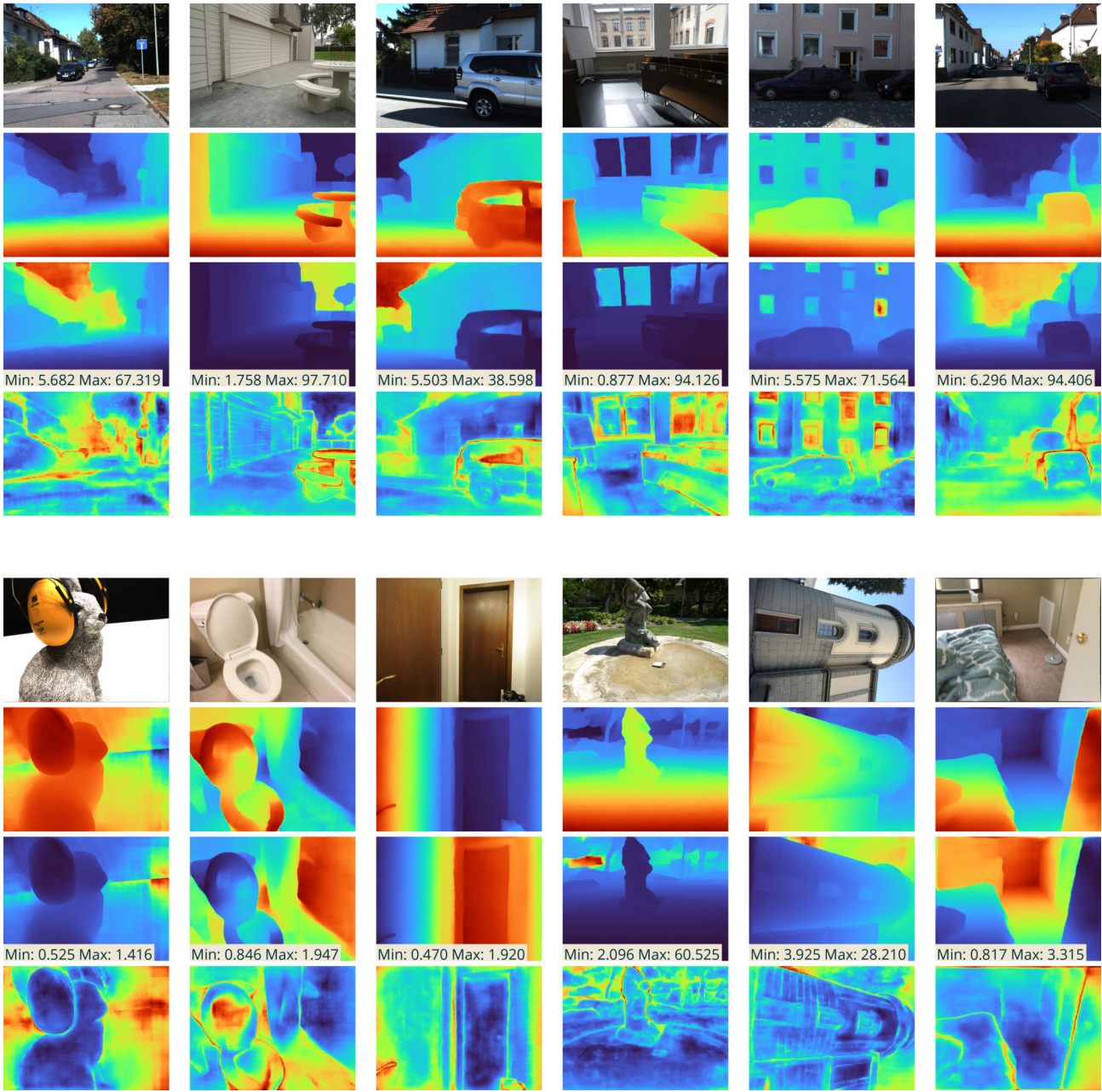

Figure A12. **Qualitative results of the Robust MVD Baseline** on samples from the KITTI, ScanNet, ETH3D, DTU, Tanks and Temples datasets. For each sample, the figure shows four images: the first row shows the keyview image, the second row the predicted inverse depth map, the third row the predicted depth map, and the fourth row the predicted uncertainty. Predicted inverse depth maps and predicted depth maps are equivalent. However, inverse depth maps are better suited for visualization, whereas depth maps allow for an easier interpretation of the scale of predicted values (here indicated in meters).


}
{}

\end{document}